\documentclass[10pt,twocolumn,letterpaper]{article}

\usepackage{wacv}
\usepackage{times}
\usepackage{epsfig}
\usepackage{graphicx}
\usepackage{amsmath}
\usepackage{amssymb}
\usepackage{booktabs}
\usepackage[accsupp]{axessibility}  

\usepackage{animate}
\usepackage{CJK}
\usepackage{xcolor}
\usepackage{url}
\usepackage{threeparttable}
\usepackage{multirow}
\usepackage[normalem]{ulem}

\def\wacvPaperID{908} 

\wacvalgorithmstrack   

\wacvfinalcopy 

\ifwacvfinal
\usepackage[breaklinks=true,bookmarks=false]{hyperref}
\else
\usepackage[pagebackref=true,breaklinks=true,colorlinks,bookmarks=false]{hyperref}
\fi

\makeatletter
\@namedef{ver@everyshi.sty}{}
\makeatother

\usepackage{tikz, pgfplots}
\pgfplotsset{compat=newest}
\usepgfplotslibrary{groupplots}

\newlength{\figheight}
\newlength{\figwidth}

\begin{document}

\title{Recur, Attend or Convolve? On Whether Temporal Modeling Matters for Cross-Domain Robustness in Action Recognition}

\author{Sofia Broomé$^1$ ~~~~ Ernest Pokropek$^1$~~~~ Boyu Li$^1$ ~~~~ Hedvig Kjellström$^{1,2}$ \\
$^1$ KTH, Sweden {\tt sbroome,pokropek,boyul,hedvig@kth.se} ~~~~ $^2$ Silo AI, Sweden}


\maketitle
\thispagestyle{empty}

\begin{figure*}[h!]
 \centering
 \begingroup
 \renewcommand{\arraystretch}{0.3}
 \begin{tabular}{ccccc}
     \rotatebox{90}{\quad \quad Less texture} & 
     \animategraphics[width=100pt]{5}{figures/diving/s1_8qRmKunCjtY_00016/mask}{0}{31} &
     \animategraphics[width=100pt]{5}{figures/diving/s2_8qRmKunCjtY_00016/mask}{0}{31} &
     \animategraphics[width=100pt]{5}{figures/diving/t_8qRmKunCjtY_00016/mask}{0}{31} &
     \rotatebox{90}{\quad \quad More texture} \\ 
     & S1 & S2 & T &  \\      
 \end{tabular}
 \endgroup
\caption{\textbf{Animated figure, displayed on click in Adobe Reader.} Example clip showing our three modified domains of Diving48 enabling the investigation of texture bias in video models: S1 (segmented divers over a blurred background), S2, (cropped bounding boxes of divers over a blurred background), and T (masked boxes of divers, and the original background).}
\label{fig:diving48}
\end{figure*}

\begin{abstract}
Most action recognition models today are highly parameterized, and evaluated on datasets with appearance-wise distinct classes. It has also been shown that 2D Convolutional Neural Networks (CNNs) tend to be biased toward texture rather than shape in still image recognition tasks \cite{geirhos2018imagenettrained}, in contrast to humans. Taken together, this raises suspicion that large video models partly learn spurious spatial texture correlations rather than to track relevant shapes over time to infer generalizable semantics from their movement. A natural way to avoid parameter explosion when learning visual patterns over time is to make use of recurrence. Biological vision consists of abundant recurrent circuitry, and is superior to computer vision in terms of domain shift generalization. In this article, we empirically study whether the choice of low-level temporal modeling has consequences for texture bias and cross-domain robustness. 
In order to enable a light-weight and systematic assessment of the ability to capture temporal structure, not revealed from single frames, we provide the \textit{Temporal Shape} (TS) dataset, as well as modified domains of Diving48 allowing for the investigation of spatial texture bias in video models. The combined results of our experiments indicate that sound physical inductive bias such as recurrence in temporal modeling may be advantageous when robustness to domain shift is important for the task. 
\end{abstract}

\section{Introduction}
\label{sec:intro}

 One of the most fundamental questions when it comes to video understanding is how to model the dependency between frames in such a way that  temporal relationships relevant to the activity in the video can be learned. A robust action recognition system should be able to figure out how frames relate to each other, and which shapes and objects have changed or persisted over time. With this knowledge, it can start to infer relationships at a higher level, such as object-object or agent-object relationships.  
  
 Three principally different approaches to frame dependency are 3D convolutions, self-attention and recurrence. 
 These methods model the world (the visual sequence) in principally different manners:  linearly, non-linearly, and non-linearly with a time-causal direction, meaning that they each use different inductive biases for the temporal modeling (\textbf{Fig. \ref{fig:rac_overview}}).  In spite of its essentiality, the frame dependency question
 has almost disappeared from action recognition articles, possibly in the race to improve on the classification benchmarks. Emphasis is instead placed on other aspects of deep video models, such as advanced architectural superstructures, regularization or training schemes. A shift toward attention-based video models has recently taken place, but without a discussion of the physical interpretation of its underlying temporal model. 
 
 Humans are still significantly stronger at generalization than artificial neural networks in vision tasks \cite{geirhos2018generalisation, Serre2019DeepLT}. Recurrent models are critical in the only visual system that has been ‘solved’ to date -- biological vision \cite{Angelucci2006ContributionOF,diLollo2000CompetitionFC,Douglas1995RecurrentEI,Douglas2007RecurrentNC,Fahrenfort2007MaskingDR,Kreiman2020BeyondTF,Lamme2000TheDM,Salin1995CorticocorticalCI,Supr2001TwoDM}. Based on the observation that feedback connections are abundant in biological but not computer vision \cite{Kreiman2020BeyondTF,KriegeskorteGoinginCircles}, in this article, we hypothesize that the lack of recurrence when learning spatiotemporal features may be one reason for this discrepancy. 
 We therefore investigate the following research question empirically in extensive and systematic experiments: does the principally different mathematical natures of 3D convolutions, self-attention and recurrence affect cross-domain robustness in video models -- and in particular, does recurrence bring about an advantage?
 

Video models lack robustness to domain shift \cite{WhyDanceChoi2019,VideoDGPAMI21,videocorruption2021}, and it has been repeatedly shown \cite{WhyDanceChoi2019, Hara2021RethinkingTD, Diving48_Li_2018_ECCV, Xie2018RethinkingSF} that the datasets most frequently cited during the 2010s (UCF-101 \cite{ucf101}, HMDB \cite{hmdbiccv2011}, Kinetics \cite{KineticsDataset}) exhibit significant spatial biases. This is a plausible reason for the poor cross-domain robustness in action recognition, since overly relying on spatial rather than motion cues intuitively results in overfitting to one domain (e.g., certain backgrounds, viewpoints or similar actor appearances).  

Contemporary state-of-the-art approaches to action recognition are predominantly either fully convolutional \cite{carreira2017quo,SlowFastFeichtenhofer_2019_ICCV,Feichtenhofer2021ALS,Ghadiyaram2019LargeScaleWP,Xie2018RethinkingSF}, combine convolutions with temporal sampling and fusion strategies \cite{Wang2021TDNTD,Wang2019TemporalSN,zhou2017temporalrelation}, or, more recently, attention-based Video Transformers (VTs) \cite{Timesformergberta_2021_ICML,VideoATGirdhar2019,videolightformerkoot2021,Selva2022VideoTransformersSurvey, vimpactan2021}. The sheer size of the models, typically more than 50M trainable parameters, gives them a strong capacity to learn in-domain patterns. As models grow larger, ever more resources are spent to train them. State-of-the-art models should display competitive benchmarking numbers on large-scale datasets, such as Kinetics-400 and Kinetics-600. It is questionable whether these benchmarks are suitable for temporal modeling, or rather for how large amounts of YouTube clips efficiently can be stored as weight representations. At the same time, the reciprocal dependency between the hardware and software of standard graphics processing units (GPUs), on the one hand, and models requiring massive parallel computation for their training, on the other hand, is becoming ever more intertwined \cite{Wired2021, NvidiaTransformersBlog}. The question looms whether we have cornered ourselves in action recognition, in the expectancy to work on ever larger models, in industry as well as in academia.
 
Theoretical works \cite{Belkin15849,Ma2018ThePO,Soltanolkotabi2019TheoreticalII} have indicated that overparametrization helps generalization, in that local minima of the loss landscape for such models often are thought to be global. These studies are made on held-out data, but never on data with significant domain shift, to the best of our knowledge.


Although less efficient to train on GPUs, recurrent video models have a more parameter-efficient approach per timestep, which may hinder over-reliance on texture cues, and promote learning the temporally relevant motion cues. 
The need to be economical with the use of trainable parameters, we hypothesize, creates incitement to learn better shape representations instead of texture representations. In turn, this allows for better generalization across datasets and in the wild. For contour detection, it was found that a model with recurrent dynamics was more sample-efficient and generalized better than a feed-forward model \cite{LinsleyNeurIPSAGLS20, LinsleyICLRKAS20}.

The primary contributions of our paper are as follows:

\vspace{-1mm}
\begin{itemize}
\setlength{\itemsep}{0pt}
    \item We present the first empirical results from systematic experiments on how the choice of frame dependency modeling in action recognition can affect cross-domain robustness.
    \item We introduce a lightweight dataset allowing for investigation of both temporal shape modeling ability and domain generalization, called the Temporal Shape dataset.
    \item We provide the first discussion and experiments on shape vs.~texture bias (following Geirhos et al.~\cite{geirhos2018imagenettrained}) in deep video models.
    \item We make segmentation-based shape and texture versions of the Diving48 dataset public (as well as 303 instance-segmented frames), allowing studies on whether a video model has learned to rely more on (temporal) shape or on texture. 
\end{itemize}
\section{Related Work}

\paragraph{Domain shift in action recognition.}

In \cite{Chen2019TemporalAA, VideoDGPAMI21}, 
cross-domain datasets are introduced to study methods for video domain adaptation. \cite{Chen2019TemporalAA} proposes to align the temporal features where the domain shift is most notable, whereas \cite{VideoDGPAMI21} proposes to improve the generalizability of so-called local features instead of global features, and use a novel augmentation scheme. Strikingly, however, all experiments in \cite{Chen2019TemporalAA, VideoDGPAMI21} are based on features extracted frame-by-frame, by a 2D ResNet \cite{ResNetHe2016}, and aggregated after-the-fact, 
meaning that they in effect do not handle spatiotemporal features. Using frame-wise features saves large amounts of time and computation, but it avoids an essential aspect of video modeling. 
Different from the field of Domain adaptation, we are not proposing methods on top of base architectures to reduce domain shift, but rather study empirically which types of fundamental video models inherently seem to be more robust to it. In an important work by Yi et al.~\cite{videocorruption2021}, benchmarks are introduced to study robustness against common video corruptions, evaluated for spatiotemporal attention- and convolution-based models. Different from our work, the domain shift is restricted to data corruptions rather than the same classification task in a new domain, and recurrent models are not evaluated.

\paragraph{Emphasis on temporality in action recognition.}

Many works emphasize the importance of temporal modeling, as the field of video understanding is growing, e.g., \cite{Dwibedi_2018_CVPR_Workshops, GhodratiBMVC2018, ManttariBroome_2020_Interpreting_Video_Features, Seeingarrowoftime14, SevillaLara2021OnlyTC, Sigurdsson2017, Xie2018RethinkingSF, zhou2017temporalrelation}. 
\cite{GhodratiBMVC2018} and \cite{ManttariBroome_2020_Interpreting_Video_Features} compare temporal modeling abilities between principally different architectures, but without explicitly investigating domain shift generalization. 
\cite{Sigurdsson2017} examines video architectures 
and datasets for human activity understanding on a number of qualitative attributes such as pose variability, brevity and density of the actions. \cite{Huang_2018_CVPR} investigates how much the motion contributes to the classification performance of 
the C3D architecture \cite{Tran2015LearningSF}. 
Both \cite{Chen_2021_CVPR} and \cite{Tran2018ACloserLook}  perform large-scale studies of the features of different variants of 2D and 3D CNNs in action recognition. Last, we are connected to \cite{SevillaLara2021OnlyTC}, which discusses the risk that models with strong image modeling abilities may prioritize those cues over the temporal modeling cues. Reminiscent of the findings of \cite{geirhos2018imagenettrained}, the authors of \cite{SevillaLara2021OnlyTC} find that inflated convolutions tend to learn classes better where motion is less important, and that generalization can be helped by training on more temporally focused data (in analogy to training on shape-based data in \cite{geirhos2018imagenettrained}). Different from our work, however, only fully convolutional models are studied and the focus is not on comparing models with fundamentally different approaches to frame dependency modeling.

\section{Experiment design}

In this section, we describe the experiment design for the two datasets: Temporal Shape and Diving48. 

\vspace{-4mm}
\paragraph{Main idea.}

In all experiments, we begin by training on a specific domain, and validating on a held-out dataset from the same domain. We save the model checkpoint which performed the best on the validation set, and then proceed to evaluate it on other domains that are different in some respects but share the same task. Following \cite{Chen2019TemporalAA}, the domain we train on will be referred to as the source, and the unseen domains that we evaluate on as the target. To measure cross-domain robustness, we define the \textbf{robustness ratio (rr.)}~ as the ratio between a model's accuracy on a target domain and its best validation accuracy on the source domain. When the target task corresponds to the source task, this number should ideally be close to one (higher is better). It can be noted that the rr. is a heuristic metric, which builds on the assumption that the performance on the in-domain validation set typically is higher than on other domains. If the performance on the validation set is poor to begin with, the rr. is less informative.

\vspace{-3mm}
\paragraph{Method common to all experiments.}
In our study, we are purposefully comparing the basic functionality of models. No pre-training, dropout, or data augmentation is applied in our experiments, except for 50\% chance of horizontal flipping of the clips on Diving48. 
Sequences are uniformly sub-sampled into equal length
(a fixed input size is required for the input to both 3D CNNs and attention-based models). 
There are 
non-uniform frame sampling methods, which can be used as augmentation, or as informed priors (e.g., the TimeSformer only samples the middle frames during inference in \cite{Timesformergberta_2021_ICML}); these are thus not used in our study, in order to study the bare bones of the models. 
Code related to neural networks was written in PyTorch \cite{pytorchNEURIPS2019_9015} using Lightning \cite{lightningfalcon2020framework}. 
Further implementation details and code can be found in the corresponding repositories ( \href{https://github.com/sofiabroome/temporal-shape-dataset}{Temporal Shape experiments}, \href{https://github.com/sofiabroome/cross-dataset-generalization}{Diving48 experiments} and \href{https://github.com/sofiabroome/diver-segmentation}{diver segmentation}). The datasets are available for download on Harvard Dataverse and linked to from the repositories. 

\subsection{Models}

We will compare ConvLSTMs, 3D CNNs and VTs, since these present three principally different temporal modeling approaches with varying types and degrees of inductive bias. As VT, we will use the TimeSformer \cite{Timesformergberta_2021_ICML}, because it recently achieved state-of-the-art results on a number of action recognition benchmarks. 

It is a challenging task to compare neural network models which have principally different architectures. In our work, we decided on controlling for three different factors: the performance on a particular dataset, the number of trainable parameter and the layer structure (i.e., the number and expressivity of hierarchical abstractions). The experiments were designed prior to running them, to keep the process as unbiased as possible. The experiments are further completely reproducible as they were run on five fixed random seeds throughout the study. 

\vspace{-2mm}
\paragraph{Convolutional LSTMs.}
The ConvLSTM \cite{Shi2015ConvolutionalLN} layer functions like an LSTM layer \cite{Hochreiter:1997:LSM:1246443.1246450}, but with matrix multiplication 
replaced with 2D convolutions. This crucially means that they allow for the input to maintain its spatial structure, contrary to classical recurrent layers 
which require a flattened input.  
Frame dependency is modeled using recurrence, which introduces non-linearities between timesteps. Further, time can only flow in the causal direction. 
A ConvLSTM video model, in this work, is a model fully based on these types of layers, with a classification head on top.

\vspace{-3mm}
\paragraph{TimeSformer.}
The TimeSformer (hereon, TimeSf) \cite{Timesformergberta_2021_ICML} is a VT, relying entirely on self-attention mechanisms to  model frame dependency. As in \cite{ViTICLR21DosovitskiyB0WZ21}, each frame is first divided into patches, which are flattened. We use the TimeSformer-PyTorch library \cite{TimesformerPytorch}, mainly with the standard settings unless otherwise specified (divided space-time attention). 
Self-attention is applied both among the patches of one frame (spatial attention) and across patches located in the same positions across the temporal axis (temporal attention). Two variants are used in the TS experiments, with the number of heads $\mathcal{A}$ set to either 1 or 8 
(TimeSf-1 and TimeSf-8). TimeSf-1 is closer to the ConvLSTM and 3D CNN in terms of parameter count, whereas TimeSf-8 is the standard setting. We note again that in order to study the fundamental behavior of the models, we do not use pre-training, advanced data augmentation, nor averaging over multiple predictions. This results in a lower performance on Diving48 for TimeSf than its state-of-the-art results. In order to control for layer structure or number of parameters which requires architectural modifications, it is not possible to use a pre-trained checkpoint. It is well-known that VTs, or Vision Transformers (ViTs) in general, require a lot of training data due to their minimal inductive bias. We therefore stress that we are not questioning the overall performance of these models -- a pre-trained version would have performed better on the Diving48 task than in our experiments, but we are investigating the fundamental behavior of models in our experiments, prior to more advanced or large-scale training schemes.   

\vspace{-3mm}
\paragraph{3D CNNs.}
In a 3D CNN, time is treated as space, and thus the input video as a volume, across which we convolve local filter volumes. Convolution is a linear operation, 
meaning that the order of frames that the 3D filter traverses does not matter. Instead, all non-linearities are applied hierarchically, 
between layers, which is how this model still can learn the arrow of time. 
Its layer structure is typically similar to a 2D CNN, including batch normalization and pooling. This is also the case for the instances used in our study.

\subsection{Experiments on the Temporal Shape dataset}

\begin{figure*}[h!]
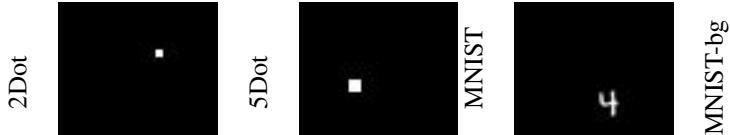

\centering
\begingroup
\renewcommand{\arraystretch}{0.3}
\begin{tabular}{cccccccc}
    \rotatebox{90}{\quad \quad 2Dot} & 
    \animategraphics[width=60pt]{5}{figures/temporal_shape_circle/2dot/12}{40}{59} & 
    \rotatebox{90}{\quad \quad 5Dot} &
    \animategraphics[width=60pt]{5}{figures/temporal_shape_circle/5dot/17}{00}{19}  
    \rotatebox{90}{\quad \quad MNIST} &
    \animategraphics[width=60pt]{5}{figures/temporal_shape_circle/14mnist_bg0/1}{40}{59} & 
    \rotatebox{90}{\quad MNIST-bg} &
    \animategraphics[width=60pt]{5}{figures/temporal_shape_circle/14mnist_bg1/15}{00}{19} \\ 
\end{tabular}
\endgroup
\caption{\textbf{The videos can be displayed on click in Adobe Reader.} Example clip showing the four domains of the TS dataset, for the class \textit{circle}. In 2Dot and 5Dot, the circle is drawn with a square of width 2 and 5 pixels. In MNIST and MNIST-bg, the circle is drawn with a MNIST digit, w/ and w/o a Perlin noise background.}
\label{fig:temporalshape}
\end{figure*}

Our proposed TS dataset is a synthetically created dataset for classification of short clips showing either a square dot or a random MNIST digit tracing shapes with their trajectories over time (Fig.~\ref{fig:temporalshape}). The dataset has five different trajectory classes (i.e., \textit{temporal shapes}): circle, line, arc, spiral and rectangle. The task is to recognize which class was drawn by the moving entity across the frames of the sequence. 
The spatial appearance of the moving object is not correlated with the class, and can thus not be employed in the recognition. In the
2Dot, 5Dot and MNIST domains, the background is black, and in MNIST-bg, the background contains white Perlin noise. The Perlin noise can be more or less fine-grained; scale is regulated by a random parameter $\sigma \in [1,10]$. 
The dataset can be thought of as a heavily scaled-down version of an action template dataset, such as 20BN-Something-something-v2 \cite{SomethingSomethingGoyal2017}, stripped of appearance cues.

The sequences consist of 20 64x64 frames, in grey scale. Each of the five classes has different amounts of possible variation in their states. The shapes can have varying starting positions, starting angles, direction, size and speeds. In  the  experiments,  4000  clips  were  used  for training  and  1000  for  validation (model selection), and 500 clips for evaluation only. The classes are sampled so as to obtain class balance. 

 Since the dataset is small, we use lightweight models. We control for layer structure by letting the compared models have three layers each of analogous blocks with the same number of hidden units in each. One block for the ConvLSTM and 3D CNN consists of a model-specific layer, 
 max pooling, 
 followed by batch normalization.  
 These two models used the same convolutional kernel sizes in all three layers ($3\times3$). For the TimeSformer, we used one TimeSformer layer as one block, and the latent dimension for each attention head, $D_h$, as the number of hidden units, since these were similar in scale.
 
 We run experiments for different numbers of hidden units per layer, $h \in \{ 2, 4, 6, 8, 10, 12, 16, 24, 32, 48\}$. 
 For each of the ten experiments of varying model sizes, we train the models on the five-class task on the source domain for 100 epochs, with ten epochs of early stopping patience, repeated under five different random seeds set from the beginning of the study. For TimeSf-1 and TimeSf-8, the maximum number of epochs is 300 (100 for early stopping) because they demand more epochs to converge than the other two types of models, due to their minimal inductive bias. We then evaluate the best model checkpoint from the source domain on different target domains with the same classification task. Experiments were conducted in two `directions', training on 2Dot and evaluating on the other domains, or training on MNIST-bg and evaluating on the other domains, since these represent two extremes on the continuum of less to more spatial noise. 

Training on the TS data is light-weight compared to real video data, and runs fast (in the minutes-range, up to an hour, for the model sizes we evaluated) on one GPU card. We train with a batch size of 64 in all TS experiments.

\subsection{Experiments on Diving48}

Diving48 \cite{Diving48_Li_2018_ECCV} is a well-known dataset for fine-grained and time-critical action recognition. It consists of 18k short clips with dives from 48 classes. Successfully classifying these dives requires temporal modeling, since one needs to keep track of the movements of the divers and their order. The dataset is varied appearance-wise, in terms of both backgrounds and viewpoints, which may contain unknown biases. 
The same competition sites can be present in both the training and test split, "to avoid biases for certain competitions", according to the authors \cite{Diving48_Li_2018_ECCV}. Instead, in our view, this in fact increases the risk for bias, since the ability to recognize a dive at an unseen site is never tested. It would have been preferable to separate competition locations entirely between training and test set. Thus, even though the dataset presents a very challenging classification task from a temporal modeling perspective, it is likely not free from spatial biases (as will be demonstrated by our experiments).

\vspace{-5mm}
\paragraph{Modified domains of Diving48.} We always train on the original dataset, but 
evaluate our trained models on slightly modified domains of the original test set. We modify the test set into three new domains: two based on shape and one based on texture (following Geirhos et al.~\cite{geirhos2018imagenettrained}, Fig.~\ref{fig:diving48}). To do this, we extend the concepts of shape and texture bias from  \cite{geirhos2018imagenettrained} to the temporal domain in the following way. In the shape domains, we blur the background and only maintain the segmented diver(s) (S1), or the divers and their bounding boxes (S2). In the texture domain (T), we conversely mask bounding boxes where the diver(s) are in each frame, and only keep the background. The masked boxes are filled with the average Imagenet \cite{Deng2009ImageNetAL} pixel value, following \cite{WhyDanceChoi2019}. The class evidence should lie only in the divers' movement; hence, the texture version should not contain any relevant signal, and the accuracy should ideally drop to random performance. In this way, we can study how different models drop in score when tested on the shape or texture domain, indicating both cross-domain robustness (for S1 and S2) and texture bias (for T).

\vspace{-4mm}
\paragraph{Instance Segmentation of Diving48.}

The segmentation of divers in Diving48 is detailed in the supplemental. We release 303 manually labeled frames with instance segmentation (single or double dives), since off-the-shelf COCO-trained \cite{Lin2014MicrosoftCC} networks fail at this task for the class Person, presumably because of the unusual shapes assumed in the air by the diver, or include people in the audience. 
\vspace{-5mm}
\paragraph{Training.}

Just as for TS, we deliberately avoid bells and whistles when training models on Diving48, to study their fundamental behavior. All three models are trained with the same SGD optimizer, cross-entropy loss, and a constant learning rate of 0.001. Each model is trained for 500 epochs maximally, with an early stopping patience of 30 epochs if the validation performance does not improve. The only data augmentation used is horizontal flipping of 50\% probability for the entire clip. The models are trained using PyTorch Lightning's \textit{ddp} parallelization scheme across eight A100 GPUs, with a batch size of 8 and a clip length of 32 uniformly sampled frames, at 224$\times$224.

Given that the purpose of our experiments is \textit{not} to optimize classification performance, we evaluate the models at different levels of performance, ranging from 30\% to 50\% accuracy. Some of the advanced state-of-the-art methods today, including pre-training and heavy data augmentation, obtain up to 80\% performance on Diving48, but when the dataset was introduced in 2018, and standard video methods were tested off-the-shelf on it, the best result was 27\% accuracy \cite{Diving48_Li_2018_ECCV}. Thus, the range of 30-50\% is reasonably well-performing, and well above random (which is at 2.1\%).

\vspace{-4mm}
\paragraph{Experiments.}

We conduct three different kinds of experiments on Diving48, 
namely control for: layer structure and performance (a-c), performance of the best performing variants (d), and number of parameters and performance (e-h). ConvLSTM has four blocks of 128 hidden ConvLSTM units each (14.3M params.) in all experiments.\\

\vspace{-2mm}
\noindent\textit{a-c. Controlling for layer structure and performance.} In this experiment, we let the models have four layers
, with $h=128$ 
in each. We again treat $D_h$ as the hidden unit analogy for TimeSf. 
We evaluate model checkpoints 
at different performance levels: 30\%, 35\%, and 38.3\% accuracy. The last 
accuracy, was chosen because it was the limiting, highest performance by the 3D CNN in this experiment. Having the same layer structure gives rise to a varying number of parameters for each type of model. Here, the 3D CNN has 10.6M params., 
and TimeSf 85M. \\

\noindent\textit{d. Controlling for performance only.}
Here, we compare models at their best performance, after hyperparameter search. Since it was not possible to train TimeSf to a higher accuracy than 39.7\% in all variants we tried, this experiment was only conducted with the 3D CNN and ConvLSTM. \footnote{A list of the variants we attempted with TimeSf is in the supplemental.}
The 3D CNN was an 11-layer VGG-style model (23.3M params.). The checkpoints used were both at exactly 50.07\% validation accuracy.\\

\noindent\textit{e-h.~Controlling for number of parameters and performance.}
Here, we have chosen models with a similar amount of trainable parameters, in this case 14M. To arrive at this number of parameters for TimeSf, its depth was reduced from 12 to 11, and $D_h$ and $D$ were halved, to 32 and 256, respectively, relative to the default model. 
The 3D CNN has six blocks with 128 units in each.

\section{Results and discussion}

Having presented the experimental design for both datasets, next, we discuss our empirical findings, first on TS, and then on Diving48 and its modified domains.

\vspace{-1mm}
\subsection{Temporal Shape}

\textbf{\textit{Condensed results:} TimeSf and ConvLSTM are more cross-domain robust than the 3D CNN in the absence of spatial texture bias.}
\vspace{-5mm}
\paragraph{Training on 2Dot.}

\begin{figure}[]
\centering
\setlength{\figwidth}{.49\textwidth}
\setlength{\figheight}{.15\textheight}
\begin{tikzpicture}

\definecolor{cyan}{RGB}{0,255,255}
\definecolor{darkslategray38}{RGB}{38,38,38}
\definecolor{green}{RGB}{0,128,0}
\definecolor{indianred1967882}{RGB}{196,78,82}
\definecolor{lavender234234242}{RGB}{234,234,242}
\definecolor{lightgray204}{RGB}{204,204,204}
\definecolor{mediumseagreen85168104}{RGB}{85,168,104}
\definecolor{mediumturquoise100181205}{RGB}{100,181,205}
\definecolor{steelblue76114176}{RGB}{76,114,176}

\begin{groupplot}[group style={group size=1 by 2}]
\nextgroupplot[
axis background/.style={fill=lavender234234242},
axis line style={white},
height=\figheight,
legend cell align={left},
legend style={
    nodes={scale=0.65, transform shape},
  fill opacity=0.8,
  at={(0.5,-1)},
  anchor=east,
  draw opacity=1,
  text opacity=1,
  draw=lightgray204,
  fill=lavender234234242
},
minor xtick={},
minor ytick={},
tick align=outside,
tick pos=left, 
width=\figwidth,
x grid style={white},
xmajorgrids,
xmajorticks=true,
xmin=-0.15, xmax=3.15,
xtick style={color=darkslategray38},
xtick={-1,0,1,2,3,4},
xtick={0,1,2,3},
xtick={0,1,2,3},
xtick={0,1,2,3},
xtick={0,1,2,3},
xticklabels={,Best val.,5dot,MNIST,MNIST-bg,},
xticklabels={Best val.,5dot,MNIST,MNIST-bg},
xticklabels={Best val.,5dot,MNIST,MNIST-bg},
xticklabels={Best val.,5dot,MNIST,MNIST-bg},
xticklabels={Best val.,5dot,MNIST,MNIST-bg},
xticklabel style = {font=\scriptsize,yshift=0.5ex},
yticklabel style = {font=\scriptsize,yshift=0.5ex},
y grid style={white},
xlabel=\textcolor{darkslategray38}{a) Training on 2Dot},
ymajorgrids,
ymajorticks=true,
ymin=13.6585, ymax=95.3115,
ytick style={color=darkslategray38},
ytick={10,20,30,40,50,60,70,80,90,100}
]
\path [draw=white, fill=red, opacity=0.18]
(axis cs:0,87.2)
--(axis cs:0,91.6)
--(axis cs:1,79.6)
--(axis cs:2,44.8)
--(axis cs:3,22.3)
--(axis cs:3,17.7)
--(axis cs:3,17.7)
--(axis cs:2,32.4)
--(axis cs:1,67.2)
--(axis cs:0,87.2)
--cycle;

\path [draw=white, fill=green, opacity=0.18]
(axis cs:0,61.1)
--(axis cs:0,89.5)
--(axis cs:1,57.4)
--(axis cs:2,51.6)
--(axis cs:3,21.3)
--(axis cs:3,17.5)
--(axis cs:3,17.5)
--(axis cs:2,31.4)
--(axis cs:1,31.4)
--(axis cs:0,61.1)
--cycle;

\path [draw=white, fill=blue, opacity=0.18]
(axis cs:0,69.53)
--(axis cs:0,81.67)
--(axis cs:1,53.8)
--(axis cs:2,42.11)
--(axis cs:3,19.63)
--(axis cs:3,17.37)
--(axis cs:3,17.37)
--(axis cs:2,31.29)
--(axis cs:1,41.2)
--(axis cs:0,69.53)
--cycle;

\path [draw=white, fill=cyan, opacity=0.18]
(axis cs:0,36.9)
--(axis cs:0,71.1)
--(axis cs:1,39.33)
--(axis cs:2,32.09)
--(axis cs:3,20.41)
--(axis cs:3,17.39)
--(axis cs:3,17.39)
--(axis cs:2,21.51)
--(axis cs:1,25.07)
--(axis cs:0,36.9)
--cycle;

\addplot [semithick, indianred1967882, mark=*, mark size=2, mark options={solid}]
table {%
0 89.4
1 73.4
2 38.6
3 20
};
\addlegendentry{3D CNN}
\addplot [semithick, mediumseagreen85168104, mark=*, mark size=2, mark options={solid}]
table {%
0 75.3
1 44.4
2 41.5
3 19.4
};
\addlegendentry{ConvLSTM}
\addplot [semithick, steelblue76114176, mark=*, mark size=2, mark options={solid}]
table {%
0 75.6
1 47.5
2 36.7
3 18.5
};
\addlegendentry{TimeSf-8}
\addplot [semithick, mediumturquoise100181205, mark=*, mark size=2, mark options={solid}]
table {%
0 54
1 32.2
2 26.8
3 18.9
};
\addlegendentry{TimeSf-1}

\nextgroupplot[
axis background/.style={fill=lavender234234242},
axis line style={white},
height=\figheight,
legend cell align={left},
legend style={nodes={scale=0.65, transform shape},
  fill opacity=0.8,
  draw opacity=1,
  text opacity=1,
  at={(1.0,0.8)},
  anchor=east,
  draw=lightgray204,
  fill=lavender234234242
},
minor xtick={},
minor ytick={},
tick align=outside,
tick pos=left, 
width=\figwidth,
x grid style={white},
xmajorgrids,
xmajorticks=true,
xmin=-0.15, xmax=3.15,
xtick style={color=darkslategray38},
xtick={-1,0,1,2,3,4},
xtick={0,1,2,3},
xtick={0,1,2,3},
xtick={0,1,2,3},
xtick={0,1,2,3},
xticklabels={,Best val.,MNIST,5Dot,2Dot,},
xticklabels={Best val.,MNIST,5Dot,2Dot},
xticklabels={Best val.,MNIST,5Dot,2Dot},
xticklabels={Best val.,MNIST,5Dot,2Dot},
xticklabels={Best val.,MNIST,5Dot,2Dot},
xticklabel style = {font=\scriptsize,yshift=0.5ex},
yticklabel style = {font=\scriptsize,yshift=0.5ex},
y grid style={white},
xlabel=\textcolor{darkslategray38}{b) Training on MNIST-bg},
ymajorgrids,
ymajorticks=true,
ymin=18.6105, ymax=100.4395,
ytick style={color=darkslategray38},
ytick={10,20,30,40,50,60,70,80,90,100,110}
]
\path [draw=white, fill=red, opacity=0.18]
(axis cs:0,94.65)
--(axis cs:0,96.15)
--(axis cs:1,96.72)
--(axis cs:2,30.66)
--(axis cs:3,26.86)
--(axis cs:3,24.34)
--(axis cs:3,24.34)
--(axis cs:2,27.94)
--(axis cs:1,94.88)
--(axis cs:0,94.65)
--cycle;

\path [draw=white, fill=green, opacity=0.18]
(axis cs:0,89.19)
--(axis cs:0,93.21)
--(axis cs:1,92.87)
--(axis cs:2,31.29)
--(axis cs:3,25.07)
--(axis cs:3,22.33)
--(axis cs:3,22.33)
--(axis cs:2,28.51)
--(axis cs:1,88.13)
--(axis cs:0,89.19)
--cycle;

\path [draw=white, fill=blue, opacity=0.18]
(axis cs:0,83.77)
--(axis cs:0,89.23)
--(axis cs:1,88.97)
--(axis cs:2,41.61)
--(axis cs:3,37.27)
--(axis cs:3,30.53)
--(axis cs:3,30.53)
--(axis cs:2,36.59)
--(axis cs:1,83.23)
--(axis cs:0,83.77)
--cycle;

\path [draw=white, fill=cyan, opacity=0.18]
(axis cs:0,65.03)
--(axis cs:0,77.37)
--(axis cs:1,76.34)
--(axis cs:2,34.97)
--(axis cs:3,31.51)
--(axis cs:3,23.89)
--(axis cs:3,23.89)
--(axis cs:2,27.23)
--(axis cs:1,63.06)
--(axis cs:0,65.03)
--cycle;

\addplot [semithick, indianred1967882, mark=*, mark size=2, mark options={solid}]
table {%
0 95.4
1 95.8
2 29.3
3 25.6
};
\addlegendentry{3D CNN}
\addplot [semithick, mediumseagreen85168104, mark=*, mark size=2, mark options={solid}]
table {%
0 91.2
1 90.5
2 29.9
3 23.7
};
\addlegendentry{ConvLSTM}
\addplot [semithick, steelblue76114176, mark=*, mark size=2, mark options={solid}]
table {%
0 86.5
1 86.1
2 39.1
3 33.9
};
\addlegendentry{TimeSf-8}
\addplot [semithick, mediumturquoise100181205, mark=*, mark size=2, mark options={solid}]
table {%
0 71.2
1 69.7
2 31.1
3 27.7
};
\addlegendentry{TimeSf-1}
\end{groupplot}

\end{tikzpicture}
\vspace{-2mm}
\caption{Average results (\% acc.) across ten trials with varying numbers of hidden units per layer, repeated five times each (thus in total, 50 runs per model). 
Plots corresponding to each model size can be found in the supplemental.}
\label{fig:plot_drops_both_ts}
\end{figure}
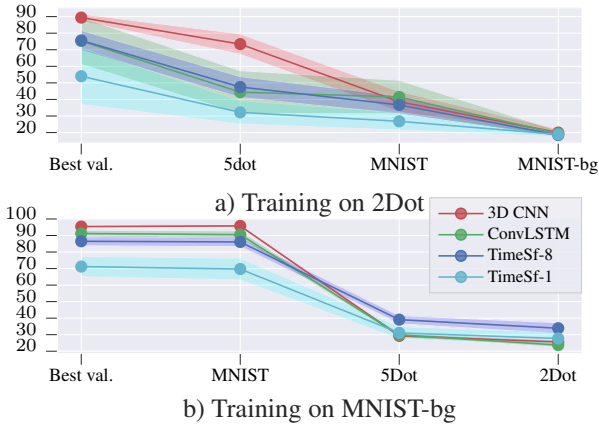

Fig.~\ref{fig:plot_drops_both_ts}a shows that although the 3D CNN generally obtains higher results on the source validation set and the nearby 5Dot domain, 
the ConvLSTM and TimeSf drop 
less compared to their original results when tested on MNIST (further from the source domain). 
ConvLSTM in fact outperforms the 3D CNN in absolute numbers on the MNIST domain. The inductive bias of a 3D CNN is highly local in space and time, which might impede learning of these temporal shapes. Generalization to the MNIST-bg domain proves too challenging for all three models.

\textit{Robustness ratio vs.~model size.} In Fig.~\ref{fig:plot_drop_vs_complexity_ts}, we have plotted the rr.~for the three target domains when training on 2Dot. For 5Dot, the rr.~for ConvLSTM decreases slightly with 
model size, whereas the 3D CNN and TimeSf, in contrast, increase the rr.~
with increased model size. 
For MNIST, which is further from the validation domain, the upward trend for the 3D CNN is broken, and less pronounced for TimeSf. For the most challenging domain, MNIST-bg, the rr.~becomes very low for all three models with increased size. The trends in Figs.~\ref{fig:plot_drop_vs_complexity_ts} a-c point to how a larger model size with promising performance in a nearby domain can potentially be an obstacle in domains that are further from the source for TimeSf and the 3D CNN. 

\begin{figure*}[t]
\centering
\setlength{\figwidth}{.33\textwidth}
\setlength{\figheight}{.15\textheight}
\input{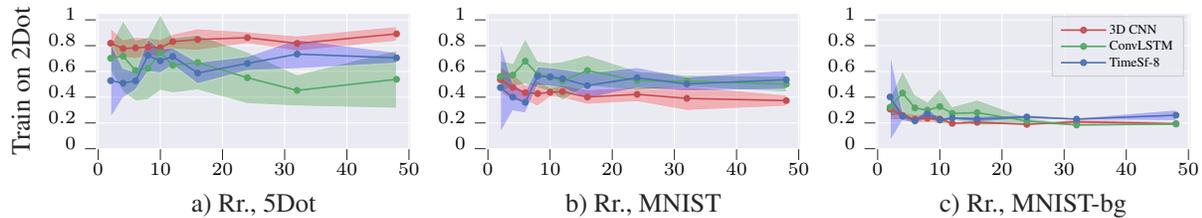}
\caption{Robustness ratio (rr.) ($\uparrow$)
when training on 2Dot, vs. number of hidden units per layer. 
The target domain is progressively further away from the source in subplots a-c. TimeSf-1 is excluded here due to its near random validation accuracy for small model sizes.}
\label{fig:plot_drop_vs_complexity_ts}
\end{figure*}

\vspace{-4mm}
\paragraph{Training on MNIST-bg.}

In this experiment, TimeSf-8 and TimeSf-1 were the most robust 
(Fig.~\ref{fig:plot_drops_both_ts} b). 
A VT is an excellent model when it comes to learning sparse, long-term dependencies in space and time. We hypothesize that this allowed TimeSf to to fully disregard the Perlin noise (which is highly stochastic and demanding to model) and 
learn the true temporal shapes, 
and that this, in turn, allowed it to be unbiased in the other domains, since the training data was designed to exclude spatial bias. 
In real-world data, however, there will always be biases, and it is therefore best to construct models which inherently encode as little bias as possible, regardless of the training data. 


\subsection{Diving48: Sensitivity to shape and texture}

\textbf{\textit{Condensed results:} ConvLSTM exhibits less texture bias and is more cross-domain robust than TimeSf and 3D CNN.}

\begin{table}[h!]
\tiny
\centering
\setlength{\tabcolsep}{3pt}
\begin{tabular}{lllll}
\textbf{Model}       & \textbf{S1/V} $\uparrow$ & \textbf{S2/V} $\uparrow$ & \textbf{T/V} $\downarrow$ & \textbf{T/S1} $\downarrow$ \\ \midrule

\textbf{3D CNN}      & $\mathbf{0.255} \pm 0.026 $   & 0.260 $\pm 0.034$  & 0.221 $\pm 0.028$         & 0.878 $\pm 0.15$ \\
\textbf{ConvLSTM}    & 0.230 $\pm 0.028$             & $\mathbf{0.266} \pm 0.026$         & $\mathbf{0.185} \pm 0.042 $   & $\mathbf{0.807} \pm 0.16$    \\
\textbf{TimeSformer} & 0.175 $\pm 0.028$            & 0.176 $\pm 0.026$             & 0.190 $\pm 0.037$ & 1.10 $\pm 0.17$    \\ \bottomrule                     
\end{tabular}
\caption{Average results for experiments a-h.}
\vspace{-4mm}
\label{table:exp-avg}
\end{table}

Table \ref{table:exp-avg} shows the average results for the Diving48 experiments. We note that ConvLSTM drops the most for T, both relative to the validation (T/V) and to the S1 (T/S1) accuracies. ConvLSTM is also most robust to the S2 domain, whereas the 3D CNN is most robust to the S1 domain. 

\begin{table}[h!]
\scriptsize
\centering
\setlength{\tabcolsep}{11.5pt}
\begin{tabular}{lllll}
\textbf{Model}       & \textbf{S1/V} $\uparrow$ & \textbf{S2/V} $\uparrow$ & \textbf{T/V} $\downarrow$ & \textbf{T/S1} $\downarrow$ \\ \midrule

\textbf{3D CNN}      & $\mathbf{0.253}$  & 0.245  & 0.257             & 1.01 \\
\textbf{ConvLSTM}    & 0.247             & $\mathbf{0.281}$             & $\mathbf{0.238}$  & $\mathbf{0.965}$      \\
\textbf{TimeSformer} & 0.198             & 0.203             & 0.250  & 1.27    \\ \bottomrule                          
\end{tabular}
\caption{Experiment a: 4x128, 30\% validation accuracy.}
\label{table:exp-a}
\end{table}

\vspace{-8mm}
\paragraph{Experiments a-d.}

\begin{figure}[t]
\centering
\setlength{\figwidth}{.25\textwidth}
\setlength{\figheight}{.21\textheight}
\begin{tikzpicture}

\definecolor{color0}{rgb}{0.917647058823529,0.917647058823529,0.949019607843137}
\definecolor{color1}{rgb}{0.768627450980392,0.305882352941176,0.32156862745098}
\definecolor{color2}{rgb}{0.333333333333333,0.658823529411765,0.407843137254902}
\definecolor{color3}{rgb}{0.298039215686275,0.447058823529412,0.690196078431373}

\begin{groupplot}[group style={group size=2 by 2}]
\nextgroupplot[
axis background/.style={fill=color0},
axis line style={white},
height=\figheight,
legend cell align={left},
legend style={  nodes={scale=0.5, transform shape}, fill opacity=0.8, draw opacity=1, text opacity=1, draw=white!80!black, fill=color0},
minor xtick={},
minor ytick={},
tick align=outside,
tick pos=left,
title={a) 4x128, 30\%},
title style={font=\scriptsize},
width=\figwidth,
x grid style={white},
xmajorgrids,
xmajorticks=true,
xmin=-0.15, xmax=3.15,
xtick style={color=white!15!black},
xtick={0,1,2,3},
xtick={0,1,2,3},
xtick={0,1,2,3},
xtick={0,1,2,3},
xticklabels={Bv.,S1,S2,T},
xticklabels={Bv.,S1,S2,T},
xticklabels={Bv.,S1,S2,T},
yticklabel style = {font=\tiny,xshift=0.5ex},
xticklabel style = {font=\tiny,xshift=0.5ex},
y grid style={white},
ymajorgrids,
ymajorticks=true,
ymin=0.04682, ymax=0.31258,
ytick style={color=white!15!black},
ytick={0,0.05,0.1,0.15,0.2,0.25,0.3,0.35}
]
\addplot [semithick, color1, mark=*, mark size=2, mark options={solid}]
table {%
0 0.3005
1 0.0761
2 0.0736
3 0.0772
};
\addlegendentry{3D CNN}
\addplot [semithick, color2, mark=*, mark size=2, mark options={solid}]
table {%
0 0.2979
1 0.0736
2 0.0838
3 0.071
};
\addlegendentry{ConvLSTM}
\addplot [semithick, color3, mark=*, mark size=2, mark options={solid}]
table {%
0 0.2979
1 0.0589
2 0.0604
3 0.0746
};
\addlegendentry{TimeSf}

\nextgroupplot[
axis background/.style={fill=color0},
axis line style={white},
height=\figheight,
legend cell align={left},
legend style={  nodes={scale=0.5, transform shape}, fill opacity=0.8, draw opacity=1, text opacity=1, draw=white!80!black, fill=color0},
minor xtick={},
minor ytick={},
tick align=outside,
tick pos=left,
title={b) 4x128, 35\% },
title style={font=\scriptsize},
width=\figwidth,
x grid style={white},
xmajorgrids,
xmajorticks=true,
xmin=-0.15, xmax=3.15,
xtick style={color=white!15!black},
xtick={0,1,2,3},
xtick={0,1,2,3},
xtick={0,1,2,3},
xtick={0,1,2,3},
xticklabels={Bv.,S1,S2,T},
xticklabels={Bv.,S1,S2,T},
xticklabels={Bv.,S1,S2,T},
yticklabel style = {font=\tiny,xshift=0.5ex},
xticklabel style = {font=\tiny,xshift=0.5ex},
y grid style={white},
ymajorgrids,
ymajorticks=true,
ymin=0.03831, ymax=0.36809,
ytick style={color=white!15!black},
ytick={0,0.05,0.1,0.15,0.2,0.25,0.3,0.35,0.4}
]
\addplot [semithick, color1, mark=*, mark size=2, mark options={solid}]
table {%
0 0.3531
1 0.0812
2 0.0822
3 0.0858
};
\addlegendentry{3D CNN}
\addplot [semithick, color2, mark=*, mark size=2, mark options={solid}]
table {%
0 0.3464
1 0.0751
2 0.094
3 0.069
};
\addlegendentry{ConvLSTM}
\addplot [semithick, color3, mark=*, mark size=2, mark options={solid}]
table {%
0 0.3497
1 0.0538
2 0.0533
3 0.0665
};
\addlegendentry{TimeSf}

\nextgroupplot[
axis background/.style={fill=color0},
axis line style={white},
height=\figheight,
legend cell align={left},
legend style={  nodes={scale=0.5, transform shape}, fill opacity=0.8, draw opacity=1, text opacity=1, draw=white!80!black, fill=color0},
minor xtick={},
minor ytick={},
tick align=outside,
tick pos=left,
title={c) 4x128, 38.3\% },
title style={font=\scriptsize},
width=\figwidth,
x grid style={white},
xmajorgrids,
xmajorticks=true,
xmin=-0.15, xmax=3.15,
xtick style={color=white!15!black},
xtick={0,1,2,3},
xtick={0,1,2,3},
xtick={0,1,2,3},
xtick={0,1,2,3},
xticklabels={Bv.,S1,S2,T},
xticklabels={Bv.,S1,S2,T},
xticklabels={Bv.,S1,S2,T},
y grid style={white},
ymajorgrids,
ymajorticks=true,
ymin=0.04322, ymax=0.39918,
ytick style={color=white!15!black},
yticklabel style = {font=\tiny,xshift=0.5ex},
xticklabel style = {font=\tiny,xshift=0.5ex},
ytick={0,0.05,0.1,0.15,0.2,0.25,0.3,0.35,0.4}
]
\addplot [semithick, color1, mark=*, mark size=2, mark options={solid}]
table {%
0 0.383
1 0.0838
2 0.084
3 0.0909
};
\addlegendentry{3D CNN}
\addplot [semithick, color2, mark=*, mark size=2, mark options={solid}]
table {%
0 0.383
1 0.107
2 0.118
3 0.0716
};
\addlegendentry{ConvLSTM}
\addplot [semithick, color3, mark=*, mark size=2, mark options={solid}]
table {%
0 0.383
1 0.0594
2 0.0594
3 0.067
};
\addlegendentry{TimeSf}

\nextgroupplot[
axis background/.style={fill=color0},
axis line style={white},
height=\figheight,
legend cell align={left},
legend style={  nodes={scale=0.5, transform shape}, fill opacity=0.8, draw opacity=1, text opacity=1, draw=white!80!black, fill=color0},
minor xtick={},
minor ytick={},
tick align=outside,
tick pos=left,
title={d) Best vars., 50\% },
title style={font=\scriptsize},
width=\figwidth,
x grid style={white},
xmajorgrids,
xmajorticks=true,
xmin=-0.15, xmax=3.15,
xtick style={color=white!15!black},
xtick={0,1,2,3},
xtick={0,1,2,3},
xtick={0,1,2,3},
xticklabels={Bv.,S1,S2,T},
xticklabels={Bv.,S1,S2,T},
yticklabel style = {font=\tiny,xshift=0.5ex},
xticklabel style = {font=\tiny,xshift=0.5ex},
y grid style={white},
ymajorgrids,
ymajorticks=true,
ymin=0.046365, ymax=0.522335,
ytick style={color=white!15!black},
ytick={0,0.1,0.2,0.3,0.4,0.5,0.6}
]
\addplot [semithick, color1, mark=*, mark size=2, mark options={solid}]
table {%
0 0.5007
1 0.117
2 0.121
3 0.0954
};
\addlegendentry{3D CNN}
\addplot [semithick, color2, mark=*, mark size=2, mark options={solid}]
table {%
0 0.5007
1 0.116
2 0.124
3 0.068
};
\addlegendentry{ConvLSTM}
\end{groupplot}

\end{tikzpicture}
\caption{Diving48 accuracy drops, from source to target, experiments a-d. Bv. is best result in the validation domain. Note how ConvLSTM drops for T, in contrast to the 3D CNN and TimeSf.}
\label{fig:plot_drop_diving_a_d}
\end{figure}
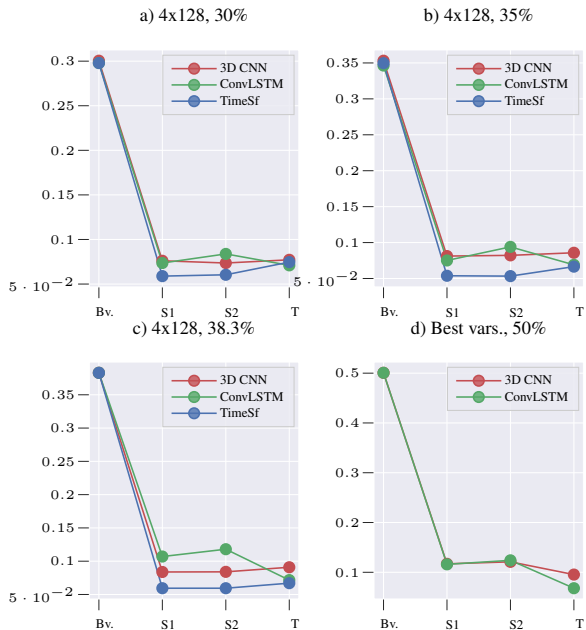

In experiments a-c (Fig.~\ref{fig:plot_drop_diving_a_d}), where we vary the validation accuracy on the source domain between 30\% and 38.3\%, 
both TimeSf and the 3D CNN perform better on T than on S1 and S2, even if only the two latter contain 
class evidence. This suggests that 
spatial bias is indeed present in Diving48, and that these models are more prone to encode it than ConvLSTM. Tables \ref{table:exp-a}-\ref{table:exp-c} show that T/S1 $>1$ 
for these two models, also visible in Fig.~\ref{fig:plot_drop_diving_a_d} a-c. \footnote{In Tables \ref{table:exp-b}-\ref{table:exp-c}, TimeSf drops the most for T relative to the validation set (T/V). This can be explained by its overall large drops, rather than being robust to texture bias, most clearly visible in Fig.~\ref{fig:plot_drop_diving_a_d} a-c. For T/V to be a meaningful metric, T/S1 should be $<1$. Therefore, we have put asterisks on the lowest T/V results which are not accompanied by T/S1 $<1$.}

\begin{table}[h!]
\scriptsize
\centering
\setlength{\tabcolsep}{11.5pt}
\begin{tabular}{lllll}
\textbf{Model}       & \textbf{S1/V} $\uparrow$ & \textbf{S2/V} $\uparrow$ & \textbf{T/V} $\downarrow$ & \textbf{T/S1} $\downarrow$ \\ \midrule

\textbf{3D CNN}      & $\mathbf{0.230}$  & 0.233  & 0.243             & 1.06 \\
\textbf{ConvLSTM}    & 0.217             & $\mathbf{0.271}$             & $\mathbf{0.199}$  & $\mathbf{0.919}$      \\
\textbf{TimeSformer} & 0.154             & 0.152             & 0.190* & 1.24    \\ \bottomrule                          
\end{tabular}
\caption{Results for experiment b: 4x128, 35\% validation accuracy. * when a low T/V result is not accompanied by T/S1 $<1$.}
\label{table:exp-b}
\end{table}


\begin{table}[h!]
\scriptsize
\centering
\setlength{\tabcolsep}{11.5pt}
\begin{tabular}{lllll}
\textbf{Model}       & \textbf{S1/V} $\uparrow$ & \textbf{S2/V} $\uparrow$ & \textbf{T/V} $\downarrow$ & \textbf{T/S1} $\downarrow$ \\ \midrule

\textbf{3D CNN}      & 0.219  & 0.219  & 0.237             & 1.09 \\
\textbf{ConvLSTM}    & $\mathbf{0.279}$             & $\mathbf{0.308}$             & $\mathbf{0.187}$  & $\mathbf{0.669}$      \\
\textbf{TimeSformer} & 0.155             & 0.155             & 0.175*  & 1.13    \\ \bottomrule                          
\end{tabular}
\caption{Results for experiment c: 4x128, 38.3\% validation accuracy. * when a low T/V result is not accompanied by T/S1 $<1$.}
\label{table:exp-c}
\end{table}

In contrast, ConvLSTM clearly drops for T. TimeSf is large here, at 85M params., whereas the 3D CNN is interestingly quite small at 10.6M params. This suggests that not only the parameter count causes susceptibility for overfitting, but that there may be innate tendencies to overfitting in the choice of spatiotemporal modeling. 
A recurrent model necessarily takes each timestep into account as it traverses
the sequence in the time-causal direction, since each timestep is non-linearly registered in the hidden state. We hypothesize that this enables it to register motion changes over time more in detail, and count these as salient, when that is the case (as it should be for Diving48).

\begin{table}[h!]
\scriptsize
\centering
\setlength{\tabcolsep}{11.5pt}
\begin{tabular}{lllll}
\textbf{Model}       & \textbf{S1/V} $\uparrow$ & \textbf{S2/V} $\uparrow$ & \textbf{T/V} $\downarrow$ & \textbf{T/S1} $\downarrow$ \\ \midrule

\textbf{3D CNN}      & $\mathbf{0.234}$   & 0.242  & 0.191             & 0.815 \\
\textbf{ConvLSTM}    & 0.232             & $\mathbf{0.248}$             & $\mathbf{0.136}$   & $\mathbf{0.586}$    \\ \bottomrule                          
\end{tabular}
\caption{Experiment d: best variants, 50.07\% val. accuracy.}
\label{table:exp-d}
\end{table}

In experiment d, where we compare a ConvLSTM and a 3D CNN at 50.07\% validation accuracy -- the best results on Diving48, the 3D CNN does not longer improve on the texture dataset relative to S1 and S2, but the drop on T is markedly larger for ConvLSTM (Table \ref{table:exp-d}).

\vspace{-4mm}
\paragraph{Qualitative examples and diving attributes.} Table \ref{tab:qual} shows a breakdown of the models' predictions on five randomly selected clips from a randomly chosen class (34). The models instances used here are from experiment c (38.3\% acc.). Top-1 acc. for these five clips being equal for all models at 0.4, we note that ConvLSTM has 100\% top-5 acc. for both S1 and S2, whereas the 3D CNN has 80\% and 60\% (40\% and 40\% for TimeSf). As for the texture (T) results, the top-5 acc. of the 3D CNN remains at 80\% relative to S1 and even improves from 60\% to 80\% relative to S2, whereas ConvLSTM drops by 40\% and TimeSf drops by 50\%. Thus, so far ConvLSTM and TimeSf display sound dropping on T. Next, we study the predictions made by the models in detail to observe that there is a qualitative difference between the predictions of ConvLSTM and TimeSf.

\begin{table}[]
\tiny
\centering
\setlength{\tabcolsep}{10pt}
\begin{tabular}{l|ll|ll|ll}
         & \multicolumn{2}{l}{\textbf{S1}} & \multicolumn{2}{l}{\textbf{S2}} & \multicolumn{2}{l}{\textbf{T}} \\ \toprule
\textbf{Model}    & Top-1   & Top-5  & Top-1   & Top-5   & Top-1  & Top-5   \\ \midrule
ConvLSTM & 0.4    & 1.0   & 0.4    & 1.0    & 0.0   & 0.6   \\
3D CNN   & 0.4    & 0.8   & 0.4    & 0.6    & 0.0   & 0.8   \\
TimeSf   & 0.4    & 0.4   & 0.2    & 0.4    & 0.0   & 0.2   \\ \bottomrule
\end{tabular}
\caption{Qualitative example with predictions on five random clips from class \textbf{34}, made by the model instances from experiment c.}
\label{tab:qual}
\end{table}

Each label of Diving48 has four attributes: takeoff, somersault, twist and flight position. Among the top-1 predictions for both S1 and S2 (Table \ref{tab:preds}), we study how many attributes are correct in the misclassifications for each model. Class 34 has the attribute values {\em inward} takeoff, {\em 2.5} somersault, {\em no} twist and {\em tuck} flight position. For ConvLSTM, the misclassifications of class 34 are 8, 20, 35 and 44, where 8, 35 and 44 all contain 3/4 correct attributes, and 20 contains 1/4 correct attributes (no twist). For the 3D CNN, only two predictions (32, 35) obtain three correct attributes, and for TimeSf, the best misclassification has only two correct attributes. 
This suggests that the 3D CNN and TimeSf have modeled the classes in terms of the true attributes to a lesser extent than ConvLSTM, i.e., ConvLSTM has learned more relevant temporal patterns, at the same global validation performance. Observing the three lower sections of Table \ref{tab:preds} for further randomly selected classes 12, 22 and 45, the ConvLSTM still achieves the largest proportion of correct attributes in the misclassifications. Just as for class 34, the 3D CNN comes second, and TimeSf last.\footnote{Tables containing the corresponding top-1 and top-5 accuracy for these additional clips are in the supplemental.}

\begin{table}[]
\tiny
\centering
\begin{tabular}{lll|ll}

         & \multicolumn{2}{l}{Top-1 predictions for five random clips from class \textbf{34}}                       &                             &                             \\ \cline{2-3}
\textbf{Model}    & \textbf{S1}            & \textbf{S2}            & \textbf{Misclassifications (set)} & \textbf{Correct attr.} \\ \toprule
ConvLSTM & {[}34, 34, 35, 8, 20{]}  & {[}34, 34, 44, 8, 20{]}  & 8, 20, 35, 44               & \textbf{10/16}                       \\
3D CNN   & {[}34, 19, 21, 35, 34{]} & {[}34, 32, 21, 21, 34{]} & 19, 21, 32, 35              & 8/16                        \\
TimeSf   & {[}34, 12, 34, 47, 20{]} & {[}31, 12, 34, 47, 20{]} & 12, 20, 31, 47              & 5/16 \\
\midrule                      
         & \multicolumn{2}{l}{Top-1 predictions for five random clips from class \textbf{12}}    &            &    \\ \cline{2-3}
ConvLSTM & {[}35, 26, 45, 26, 21{]}  & {[}27, 26, 45, 14, 21{]} & 14, 21, 26, 27, 35, 45             & \textbf{14/24} \\
3D CNN   & {[}3, 20, 12, 5, 44{]} & {[}3, 20, 12, 5, 34{]}  &  3, 5, 20, 34, 44            & 8/20 \\
TimeSf   & {[}22, 33, 12, 31, 14{]} & {[}22, 33, 12, 31, 14{]}  &  14, 22, 31, 33            & 5/16 \\ \midrule
         & \multicolumn{2}{l}{Top-1 predictions for five random clips from class \textbf{22}}    &            &    \\ \cline{2-3}
ConvLSTM & {[}26, 26, 35, 22, 21{]}  & {[}26, 26, 35, 22, 21{]} &  21, 26, 35            & \textbf{5/12} \\
3D CNN   & {[}29, 7, 26, 28, 0{]} & {[}29, 22, 26, 26, 0{]}  & 0, 7, 26, 28, 29             & 7/20 \\
TimeSf   & {[}15, 27, 46, 44, 34{]} & {[}15, 27, 46, 44, 34{]}  & 15, 27, 34, 44, 46    & 5/20 \\ \midrule
         & \multicolumn{2}{l}{Top-1 predictions for five random clips from class \textbf{45}}    &            &    \\ \cline{2-3}
ConvLSTM & {[}26, 21, 12, 35, 27{]}  & {[}26, 21, 12, 35, 44{]} &  12, 21, 26, 27, 35, 44  & \textbf{14/24} \\
3D CNN   & {[}46, 20, 35, 35, 34{]} & {[}34, 20, 12, 35, 31{]}  & 12, 20, 31, 34, 35, 46 & 12/24 \\
TimeSf   & {[}15, 31, 44, 12, 18{]} & {[}42, 31, 44, 12, 8{]}  & 8, 12, 15, 18, 31, 42, 44     & 11/28 \\
\bottomrule    
\end{tabular}
\caption{ Examples of predictions and misclassifications. 
Each class has four attributes, and the Correct attr. column shows how many attributes were correct among the set of misclassifications. }
\label{tab:preds}
\end{table}

\vspace{-5mm}
\paragraph{Experiments e-h.}
The results for experiments e-h, where the number of trainable parameters and performance are fixed, are shown in Fig.~\ref{fig:plot_drop_diving_e_h} (tabulated results in the supplemental). Here, the 3D CNN is the most robust out of the three, although ConvLSTM approaches the 3D CNN and drops more steeply for T in g-h, where the performance is higher (40\% and 45\% acc.).
In these experiments, although 
least robust, TimeSf does not improve on T relative to S1 and S2 any more. 
This suggests that TimeSf is more likely to display texture bias when it has a larger amount of parameters, as it does in experiments a-c.



\begin{figure}[]
\centering
\setlength{\figwidth}{.25\textwidth}
\setlength{\figheight}{.21\textheight}
\begin{tikzpicture}

\definecolor{color0}{rgb}{0.9176470588229,0.9176470588229,0.949019607843137}
\definecolor{color1}{rgb}{0.768627450980392,0.3058822941176,0.326862745098}
\definecolor{color2}{rgb}{0.333333333333333,0.6588229411765,0.407843137254902}
\definecolor{color3}{rgb}{0.29803926862,0.4470588229412,0.690196078431373}

\begin{groupplot}[group style={group size=2 by 2}]
\nextgroupplot[
axis background/.style={fill=color0},
axis line style={white},
height=\figheight,
legend cell align={left},
legend style={ nodes={scale=0.5, transform shape}, fill opacity=0.8, draw opacity=1, text opacity=1, draw=white!80!black, fill=color0},
minor xtick={},
minor ytick={},
tick align=outside,
tick pos=left,
title={e) 14M p., 30\% },
title style={font=\scriptsize},
width=\figwidth,
x grid style={white},
xmajorgrids,
xmajorticks=true,
xmin=-0.15, xmax=3.15,
xtick style={color=white!15!black},
xtick={0,1,2,3},
xtick={0,1,2,3},
xtick={0,1,2,3},
xtick={0,1,2,3},
xticklabels={Bv.,S1,S2,T},
xticklabels={Bv.,S1,S2,T},
xticklabels={Bv.,S1,S2,T},
yticklabel style = {font=\tiny,xshift=0.5ex},
xticklabel style = {font=\tiny,xshift=0.5ex},
y grid style={white},
ymajorgrids,
ymajorticks=true,
ymin=0.0422, ymax=0.315625,
ytick style={color=white!15!black},
ytick={0,0.05,0.1,0.15,0.2,0.25,0.3,0.35}
]
\addplot [semithick, color1, mark=*, mark size=2, mark options={solid}]
table {%
0 0.3032
1 0.0899
2 0.0959
3 0.0741
};
\addlegendentry{3D CNN}
\addplot [semithick, color2, mark=*, mark size=2, mark options={solid}]
table {%
0 0.2979
1 0.0729
2 0.083
3 0.0709
};
\addlegendentry{ConvLSTM}
\addplot [semithick, color3, mark=*, mark size=2, mark options={solid}]
table {%
0 0.2999
1 0.0633
2 0.0617
3 0.0547
};
\addlegendentry{TimeSf}

\nextgroupplot[
axis background/.style={fill=color0},
axis line style={white},
height=\figheight,
legend cell align={left},
legend style={ nodes={scale=0.5, transform shape}, fill opacity=0.8, draw opacity=1, text opacity=1, draw=white!80!black, fill=color0},
minor xtick={},
minor ytick={},
tick align=outside,
tick pos=left,
title={f) 14M p., 35\% },
title style={font=\scriptsize},
width=\figwidth,
x grid style={white},
xmajorgrids,
xmajorticks=true,
xmin=-0.15, xmax=3.15,
xtick style={color=white!15!black},
xtick={0,1,2,3},
xtick={0,1,2,3},
xtick={0,1,2,3},
xtick={0,1,2,3},
xticklabels={Bv.,S1,S2,T},
xticklabels={Bv.,S1,S2,T},
xticklabels={Bv.,S1,S2,T},
yticklabel style = {font=\tiny,xshift=0.5ex},
xticklabel style = {font=\tiny,xshift=0.5ex},
y grid style={white},
ymajorgrids,
ymajorticks=true,
ymin=0.037645, ymax=0.366655,
ytick style={color=white!15!black},
ytick={0,0.05,0.1,0.15,0.2,0.25,0.3,0.35,0.4}
]
\addplot [semithick, color1, mark=*, mark size=2, mark options={solid}]
table {%
0 0.3517
1 0.098
2 0.1066
3 0.0761
};
\addlegendentry{3D CNN}
\addplot [semithick, color2, mark=*, mark size=2, mark options={solid}]
table {%
0 0.3464
1 0.0749
2 0.0936
3 0.0688
};
\addlegendentry{ConvLSTM}
\addplot [semithick, color3, mark=*, mark size=2, mark options={solid}]
table {%
0 0.3497
1 0.0541
2 0.0567
3 0.0526
};
\addlegendentry{TimeSf}

\nextgroupplot[
axis background/.style={fill=color0},
axis line style={white},
height=\figheight,
legend cell align={left},
legend style={ nodes={scale=0.5, transform shape}, fill opacity=0.8, draw opacity=1, text opacity=1, draw=white!80!black, fill=color0},
minor xtick={},
minor ytick={},
tick align=outside,
tick pos=left,
title={g) 14M p., 40\% },
title style={font=\scriptsize},
width=\figwidth,
x grid style={white},
xmajorgrids,
xmajorticks=true,
xmin=-0.15, xmax=3.15,
xtick style={color=white!15!black},
xtick={0,1,2,3},
xtick={0,1,2,3},
xtick={0,1,2,3},
xticklabels={Bv.,S1,S2,T},
xticklabels={Bv.,S1,S2,T},
yticklabel style = {font=\tiny,xshift=0.5ex},
xticklabel style = {font=\tiny,xshift=0.5ex},
y grid style={white},
ymajorgrids,
ymajorticks=true,
ymin=0.04024, ymax=0.42436,
ytick style={color=white!15!black},
ytick={0,0.05,0.1,0.15,0.2,0.25,0.3,0.35,0.4,0.45}
]
\addplot [semithick, color1, mark=*, mark size=2, mark options={solid}]
table {%
0 0.4062
1 0.105
2 0.107
3 0.0774
};
\addlegendentry{3D CNN}
\addplot [semithick, color2, mark=*, mark size=2, mark options={solid}]
table {%
0 0.4069
1 0.0744
2 0.0911
3 0.0577
};
\addlegendentry{ConvLSTM}

\nextgroupplot[
axis background/.style={fill=color0},
axis line style={white},
height=\figheight,
legend cell align={left},
legend style={ nodes={scale=0.5, transform shape}, fill opacity=0.8, draw opacity=1, text opacity=1, draw=white!80!black, fill=color0},
minor xtick={},
minor ytick={},
tick align=outside,
tick pos=left,
title={h) 14M p., 45\% },
title style={font=\scriptsize},
width=\figwidth,
x grid style={white},
xmajorgrids,
xmajorticks=true,
xmin=-0.15, xmax=3.15,
xtick style={color=white!15!black},
xtick={0,1,2,3},
xtick={0,1,2,3},
xtick={0,1,2,3},
xticklabels={Bv.,S1,S2,T},
xticklabels={Bv.,S1,S2,T},
yticklabel style = {font=\tiny,xshift=0.5ex},
xticklabel style = {font=\tiny,xshift=0.5ex},
y grid style={white},
ymajorgrids,
ymajorticks=true,
ymin=0.0454, ymax=0.4722,
ytick style={color=white!15!black},
ytick={0,0.05,0.1,0.15,0.2,0.25,0.3,0.35,0.4,0.45,0.5}
]
\addplot [semithick, color1, mark=*, mark size=2, mark options={solid}]
table {%
0 0.4495
1 0.12
2 0.116
3 0.0855
};
\addlegendentry{3D CNN}
\addplot [semithick, color2, mark=*, mark size=2, mark options={solid}]
table {%
0 0.4528
1 0.0992
2 0.111
3 0.0648
};
\addlegendentry{ConvLSTM}
\end{groupplot}

\end{tikzpicture}
\caption{Experiments e-h on Diving48, reads as Fig. \ref{fig:plot_drop_diving_a_d}.
\vspace{-15mm}
}
\label{fig:plot_drop_diving_e_h}
\end{figure}
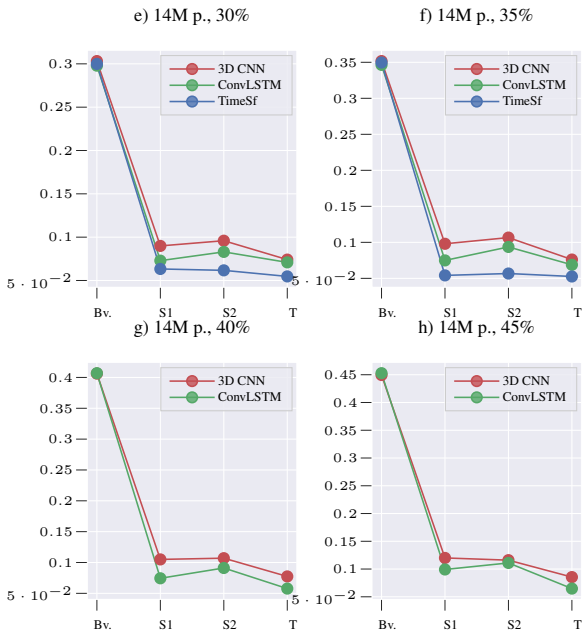


\section{Conclusions and discussion}

We have studied cross-domain robustness for three models that are principally different in terms of temporal modeling, in their bare-bones settings. A 3D CNN treats frames as a linear volume, a VT lets frames have non-linear but time-symmetric relationships, and a ConvLSTM models frame dependency non-linearly in a time-causal direction. Recently, a discrepancy in terms of feedback connections between biological and computer vision has been discussed \cite{Kreiman2020BeyondTF, KriegeskorteGoinginCircles}, and in our work we have hypothesized that the lack of feedback connections is one reason for the similarly lacking generalization abilities in computer vision. 

Our experiments were carried out on two very different datasets, one synthetic, without bias, and one with natural data, thus with more noise and potential spatial bias. 
The combined results (Figs. \ref{fig:plot_drops_both_ts}-\ref{fig:plot_drop_diving_a_d}, Tables \ref{table:exp-avg} and \ref{tab:preds}) on these datasets indicated that convolutional-recurrent temporal modeling is more robust to domain shift than self-attention and 3D convolutions in terms of bare-bones behavior, presumably owing to its lesser encoding of texture bias.  Our results are fully reproducible with public seeds, code and data. The fact that our observations regarding texture bias are made for a fine-grained dataset such as Diving48, 
constructed to contain as little bias as possible, suggests that the issue may be worse when it comes to more spatially biased datasets such as Kinetics, which is left for future work. It is furthermore left for future work whether ImageNet pre-trained VTs display more or less texture bias than their trained-from-scratch counterparts. Another observation from our study is that  when the parameter count was kept equal (experiments e-h), these trends were less pronounced.

Moreover, qualitative random examples consistently showed that the ConvLSTM learned more relevant diving patterns than the two others, when scrutinizing the three models' misclassifications -- which emphasizes the texture bias tendency of TimeSf and the 3D CNN. 
Sharing parameters across timesteps, as recurrent models do, narrows the parameter space, possibly incentivizing these models to prioritize which patterns to learn. Another reason to use smaller models is that they require less data to train, which is ethically desirable, both in that the data can be inspected more easily, and from a sustainability perspective \cite{Bender2021OnTD}. 

Our study indicates that sound physical inductive bias such as recurrence in temporal modeling may be advantageous when robustness to domain shift is important for the task. In action recognition, benchmarking has thus far mainly been conducted for in-domain-tasks where large models perform well. We encourage the video understanding community to increasingly conduct evaluation on tasks involving domain shift. We hope that our proposed datasets and framework for evaluation can help such future domain shift robustness investigations of spatiotemporal features. 

\paragraph{Acknowledgements.}
The computations were enabled by the supercomputing resource Berzelius provided by National Supercomputer Centre at Linköping University and the Knut and Alice Wallenberg foundation. We further thank Marcus Klasson, Taras Kucherenko and Ci Li for helpful feedback and discussions.

{\small
\bibliographystyle{ieee_fullname}
\bibliography{egbib}

\begin{thebibliography}{10}\itemsep=-1pt

\bibitem{Angelucci2006ContributionOF}
Alessandra Angelucci and Paul~C. Bressloff.
\newblock {Contribution of feedforward, lateral and feedback connections to the
  classical receptive field center and extra-classical receptive field surround
  of primate V1 neurons.}
\newblock {\em Progress in Brain Research}, 154:93--120, 2006.

\bibitem{Belkin15849}
Mikhail Belkin, Daniel Hsu, Siyuan Ma, and Soumik Mandal.
\newblock Reconciling modern machine-learning practice and the classical
  bias{\textendash}variance trade-off.
\newblock {\em Proceedings of the National Academy of Sciences},
  116(32):15849--15854, 2019.

\bibitem{Bender2021OnTD}
Emily~M. Bender, Timnit Gebru, Angelina McMillan-Major, and Shmargaret
  Shmitchell.
\newblock On the dangers of stochastic parrots: Can language models be too big?
\newblock In {\em ACM Conference on Fairness, Accountability, and
  Transparency}, 2021.

\bibitem{Timesformergberta_2021_ICML}
Gedas Bertasius, Heng Wang, and Lorenzo Torresani.
\newblock Is space-time attention all you need for video understanding?
\newblock In {\em International Conference on Machine Learning}, 2021.

\bibitem{carreira2017quo}
Joao Carreira and Andrew Zisserman.
\newblock {Quo Vadis, Action Recognition? A} new model and the kinetics
  dataset.
\newblock In {\em IEEE Conference on Computer Vision and Pattern Recognition},
  2017.

\bibitem{Chen_2021_CVPR}
Chun-Fu~Richard Chen, Rameswar Panda, Kandan Ramakrishnan, Rogerio Feris, John
  Cohn, Aude Oliva, and Quanfu Fan.
\newblock Deep analysis of {CNN}-based spatio-temporal representations for
  action recognition.
\newblock In {\em IEEE Conference on Computer Vision and Pattern Recognition},
  2021.

\bibitem{Chen2019TemporalAA}
Min-Hung Chen, Zsolt Kira, Ghassan Al-Regib, Jaekwon Yoo, Ruxin Chen, and Jian
  Zheng.
\newblock Temporal attentive alignment for large-scale video domain adaptation.
\newblock In {\em 2019 IEEE International Conference on Computer Vision}, 2019.

\bibitem{WhyDanceChoi2019}
Jinwoo Choi, Chen Gao, Joseph~C.E. Messou, and Jia-Bin Huang.
\newblock Why can't {I} dance in the mall? {Learning} to mitigate scene bias in
  action recognition.
\newblock In {\em Advances in Neural Information Processing Systems}, 2019.

\bibitem{Deng2009ImageNetAL}
Jia Deng, Wei Dong, Richard Socher, Li-Jia Li, K. Li, and Li Fei-Fei.
\newblock Imagenet: A large-scale hierarchical image database.
\newblock In {\em IEEE Conference on Computer Vision and Pattern Recognition},
  2009.

\bibitem{diLollo2000CompetitionFC}
Vincent di Lollo, James~T. Enns, and Ronald~A. Rensink.
\newblock Competition for consciousness among visual events: the psychophysics
  of reentrant visual processes.
\newblock {\em Journal of Experimental Psychology. General}, 129 4:481--507,
  2000.

\bibitem{ViTICLR21DosovitskiyB0WZ21}
Alexey Dosovitskiy, Lucas Beyer, Alexander Kolesnikov, Dirk Weissenborn,
  Xiaohua Zhai, Thomas Unterthiner, Mostafa Dehghani, Matthias Minderer, Georg
  Heigold, Sylvain Gelly, Jakob Uszkoreit, and Neil Houlsby.
\newblock An image is worth 16x16 words: Transformers for image recognition at
  scale.
\newblock In {\em International Conference on Learning Representations}, 2021.

\bibitem{Douglas1995RecurrentEI}
R Douglas, Christof Koch, Misha~A. Mahowald, KA Martin, and H.~E.~Ortiz Suarez.
\newblock Recurrent excitation in neocortical circuits.
\newblock {\em Science}, 269:981 -- 985, 1995.

\bibitem{Douglas2007RecurrentNC}
Rodney~J. Douglas and Kevan A.~C. Martin.
\newblock Recurrent neuronal circuits in the neocortex.
\newblock {\em Current Biology}, 17:R496--R500, 2007.

\bibitem{Dwibedi_2018_CVPR_Workshops}
Debidatta Dwibedi, Pierre Sermanet, and Jonathan Tompson.
\newblock Temporal reasoning in videos using convolutional gated recurrent
  units.
\newblock In {\em IEEE Conference on Computer Vision and Pattern Recognition
  Workshops}, 2018.

\bibitem{Fahrenfort2007MaskingDR}
Johannes~Jacobus Fahrenfort, H.~Steven Scholte, and Victor A.~F. Lamme.
\newblock Masking disrupts reentrant processing in human visual cortex.
\newblock {\em Journal of Cognitive Neuroscience}, 19:1488--1497, 2007.

\bibitem{lightningfalcon2020framework}
William Falcon and Kyunghyun Cho.
\newblock A framework for contrastive self-supervised learning and designing a
  new approach.
\newblock {\em arXiv preprint arXiv:2009.00104}, 2020.

\bibitem{SlowFastFeichtenhofer_2019_ICCV}
Christoph Feichtenhofer, Haoqi Fan, Jitendra Malik, and Kaiming He.
\newblock {SlowFast} networks for video recognition.
\newblock In {\em IEEE International Conference on Computer Vision}, 2019.

\bibitem{Feichtenhofer2021ALS}
Christoph Feichtenhofer, Haoqi Fan, Bo Xiong, Ross~B. Girshick, and Kaiming He.
\newblock A large-scale study on unsupervised spatiotemporal representation
  learning.
\newblock In {\em IEEE Conference on Computer Vision and Pattern Recognition},
  2021.

\bibitem{geirhos2018imagenettrained}
Robert Geirhos, Patricia Rubisch, Claudio Michaelis, Matthias Bethge, Felix~A.
  Wichmann, and Wieland Brendel.
\newblock Imagenet-trained {CNN}s are biased towards texture; increasing shape
  bias improves accuracy and robustness.
\newblock In {\em International Conference on Learning Representations}, 2019.

\bibitem{geirhos2018generalisation}
Robert Geirhos, Carlos~RM Temme, Jonas Rauber, Heiko~H Sch{\"u}tt, Matthias
  Bethge, and Felix~A Wichmann.
\newblock Generalisation in humans and deep neural networks.
\newblock In {\em Advances in Neural Information Processing Systems}, 2018.

\bibitem{Ghadiyaram2019LargeScaleWP}
Deepti Ghadiyaram, Matt Feiszli, Du Tran, Xueting Yan, Heng Wang, and
  Dhruv~Kumar Mahajan.
\newblock Large-scale weakly-supervised pre-training for video action
  recognition.
\newblock In {\em IEEE Conference on Computer Vision and Pattern Recognition},
  2019.

\bibitem{GhodratiBMVC2018}
A. Ghodrati, E. Gavves, and C.~G.~M. Snoek.
\newblock Video time: Properties, encoders and evaluation.
\newblock In {\em British Machine Vision Conference}, 2018.

\bibitem{VideoATGirdhar2019}
Rohit Girdhar, Jo{\~a}o Carreira, Carl Doersch, and Andrew Zisserman.
\newblock Video action transformer network.
\newblock In {\em IEEE Conference on Computer Vision and Pattern Recognition},
  2019.

\bibitem{SomethingSomethingGoyal2017}
Raghav Goyal, Samira~Ebrahimi Kahou, Vincent Michalski, Joanna Materzynska,
  Susanne Westphal, Heuna Kim, Valentin Haenel, Ingo Fr{\"u}nd, Peter~N.
  Yianilos, Moritz Mueller-Freitag, Florian Hoppe, Christian Thurau, Ingo Bax,
  and Roland Memisevic.
\newblock The “something something” video database for learning and
  evaluating visual common sense.
\newblock In {\em IEEE International Conference on Computer Vision}, 2017.

\bibitem{Hara2021RethinkingTD}
Kensho Hara, Yuchi Ishikawa, and Hirokatsu Kataoka.
\newblock Rethinking training data for mitigating representation biases in
  action recognition.
\newblock In {\em IEEE Conference on Computer Vision and Pattern Recognition
  Workshops}, 2021.

\bibitem{ResNetHe2016}
Kaiming He, X. Zhang, Shaoqing Ren, and Jian Sun.
\newblock Deep residual learning for image recognition.
\newblock In {\em IEEE Conference on Computer Vision and Pattern Recognition},
  2016.

\bibitem{Hochreiter:1997:LSM:1246443.1246450}
Sepp Hochreiter and J\"{u}rgen Schmidhuber.
\newblock Long short-term memory.
\newblock {\em Neural Comput.}, 9(8):1735--1780, Nov. 1997.

\bibitem{Huang_2018_CVPR}
De-An Huang, Vignesh Ramanathan, Dhruv Mahajan, Lorenzo Torresani, Manohar
  Paluri, Li Fei-Fei, and Juan Carlos~Niebles.
\newblock What makes a video a video: Analyzing temporal information in video
  understanding models and datasets.
\newblock In {\em IEEE Conference on Computer Vision and Pattern Recognition},
  2018.

\bibitem{KineticsDataset}
Will Kay, Jo{\~{a}}o Carreira, Karen Simonyan, Brian Zhang, Chloe Hillier,
  Sudheendra Vijayanarasimhan, Fabio Viola, Tim Green, Trevor Back, Paul
  Natsev, Mustafa Suleyman, and Andrew Zisserman.
\newblock The kinetics human action video dataset.
\newblock {\em CoRR}, abs/1705.06950, 2017.

\bibitem{Wired2021}
Nicole Kobie.
\newblock {NVIDIA and the battle for the future of AI chips}.
\newblock {\em Wired}, 2021.

\bibitem{videolightformerkoot2021}
Raivo Koot and Haiping Lu.
\newblock Videolightformer: Lightweight action recognition using transformers.
\newblock {\em arXiv preprint arXiv:2107.00451}, 2021.

\bibitem{Kreiman2020BeyondTF}
G. Kreiman and Thomas Serre.
\newblock Beyond the feedforward sweep: feedback computations in the visual
  cortex.
\newblock {\em Annals of the New York Academy of Sciences}, 1464, 2020.

\bibitem{hmdbiccv2011}
H. Kuehne, H. Jhuang, E. Garrote, T. Poggio, and T. Serre.
\newblock {HMDB: A} large video database for human motion recognition.
\newblock In {\em IEEE International Conference on Computer Vision}, 2011.

\bibitem{Lamme2000TheDM}
Victor A.~F. Lamme and Pieter~R. Roelfsema.
\newblock The distinct modes of vision offered by feedforward and recurrent
  processing.
\newblock {\em Trends in Neurosciences}, 23:571--579, 2000.

\bibitem{Diving48_Li_2018_ECCV}
Yingwei Li, Yi Li, and Nuno Vasconcelos.
\newblock {RESOUND: Towards} action recognition without representation bias.
\newblock In {\em European Conference on Computer Vision}, 2018.

\bibitem{Lin2014MicrosoftCC}
Tsung-Yi Lin, Michael Maire, Serge~J. Belongie, James Hays, Pietro Perona, Deva
  Ramanan, Piotr Doll{\'a}r, and C.~Lawrence Zitnick.
\newblock {Microsoft COCO: Common Objects in Context}.
\newblock In {\em European Conference on Computer Vision}, 2014.

\bibitem{LinsleyNeurIPSAGLS20}
Drew Linsley, Alekh~Karkada Ashok, Lakshmi~Narasimhan Govindarajan, Rex Liu,
  and Thomas Serre.
\newblock Stable and expressive recurrent vision models.
\newblock In Hugo Larochelle, Marc'Aurelio Ranzato, Raia Hadsell,
  Maria{-}Florina Balcan, and Hsuan{-}Tien Lin, editors, {\em Advances in
  Neural Information Processing Systems}, 2020.

\bibitem{LinsleyICLRKAS20}
Drew Linsley, Junkyung Kim, Alekh Ashok, and Thomas Serre.
\newblock Recurrent neural circuits for contour detection.
\newblock In {\em International Conference on Learning Representations}, 2020.

\bibitem{Ma2018ThePO}
Siyuan Ma, Raef Bassily, and Mikhail Belkin.
\newblock {The Power of Interpolation: Understanding the Effectiveness of SGD
  in Modern Over-parametrized Learning}.
\newblock In {\em International Conference on Machine Learning}, 2018.

\bibitem{ManttariBroome_2020_Interpreting_Video_Features}
Joonatan M\"antt\"ari*, Sofia Broom\'e*, John Folkesson, and Hedvig
  Kjellstr\"om.
\newblock {Interpreting Video Features: a Comparison of 3D Convolutional
  Networks and Convolutional LSTM Networks}.
\newblock In {\em Asian Conference on Computer Vision. (*Joint first authors)},
  2020.

\bibitem{NvidiaTransformersBlog}
Alexandre Milesi.
\newblock {Accelerating SE(3)-Transformers Training Using an NVIDIA Open-Source
  Model Implementation}.
\newblock \url{https://bit.ly/3wQac3v/}.
\newblock Accessed: 2021-11-01.

\bibitem{pytorchNEURIPS2019_9015}
Adam Paszke, Sam Gross, Francisco Massa, Adam Lerer, James Bradbury, Gregory
  Chanan, Trevor Killeen, Zeming Lin, Natalia Gimelshein, Luca Antiga, Alban
  Desmaison, Andreas Kopf, Edward Yang, Zachary DeVito, Martin Raison, Alykhan
  Tejani, Sasank Chilamkurthy, Benoit Steiner, Lu Fang, Junjie Bai, and Soumith
  Chintala.
\newblock Pytorch: An imperative style, high-performance deep learning library.
\newblock In {\em Advances in Neural Information Processing Systems}, 2019.

\bibitem{Seeingarrowoftime14}
Lyndsey~C. Pickup, Zheng Pan, Donglai Wei, YiChang Shih, Changshui Zhang,
  Andrew Zisserman, Bernhard Sch{\"o}lkopf, and William~T. Freeman.
\newblock Seeing the arrow of time.
\newblock In {\em IEEE Conference on Computer Vision and Pattern Recognition},
  2014.

\bibitem{Salin1995CorticocorticalCI}
Paul~Antoine Salin and J. Bullier.
\newblock Corticocortical connections in the visual system: structure and
  function.
\newblock {\em Physiological Reviews}, 75 1:107--54, 1995.

\bibitem{Selva2022VideoTransformersSurvey}
Javier Selva, Anders~S. Johansen, Sergio Escalera, Kamal Nasrollahi, Thomas~B.
  Moeslund, and Albert Clapés.
\newblock Video transformers: A survey.
\newblock {\em arXiv preprint arXiv:2201.05991}, 2022.

\bibitem{Serre2019DeepLT}
Thomas Serre.
\newblock Deep learning: The good, the bad, and the ugly.
\newblock {\em Annual Review of Vision Science}, 2019.

\bibitem{SevillaLara2021OnlyTC}
Laura Sevilla-Lara, Shengxin Zha, Zhicheng Yan, Vedanuj Goswami, Matt Feiszli,
  and Lorenzo Torresani.
\newblock Only time can tell: Discovering temporal data for temporal modeling.
\newblock In {\em IEEE Winter Conference on Applications of Computer Vision},
  2021.

\bibitem{Shi2015ConvolutionalLN}
Xingjian Shi, Zhourong Chen, Hao Wang, Dit-Yan Yeung, Wai-Kin Wong, and Wang
  chun Woo.
\newblock {Convolutional LSTM Network: A} machine learning approach for
  precipitation nowcasting.
\newblock In {\em Advances in Neural Information Processing Systems}, 2015.

\bibitem{Sigurdsson2017}
Gunnar~A. Sigurdsson, Olga Russakovsky, and Abhinav~Kumar Gupta.
\newblock What actions are needed for understanding human actions in videos?
\newblock In {\em IEEE International Conference on Computer Vision}, 2017.

\bibitem{Soltanolkotabi2019TheoreticalII}
Mahdi Soltanolkotabi, Adel Javanmard, and J. Lee.
\newblock Theoretical insights into the optimization landscape of
  over-parameterized shallow neural networks.
\newblock {\em IEEE Transactions on Information Theory}, 65:742--769, 2019.

\bibitem{ucf101}
Khurram Soomro, Amir~Roshan Zamir, and Mubarak Shah.
\newblock {UCF101: A Dataset of 101 Human Actions Classes From Videos in The
  Wild}.
\newblock {\em CoRR}, abs/1212.0402, 2012.

\bibitem{Supr2001TwoDM}
Hans Sup{\`e}r, Henk Spekreijse, and Victor A.~F. Lamme.
\newblock Two distinct modes of sensory processing observed in monkey primary
  visual cortex (v1).
\newblock {\em Nature Neuroscience}, 4:304--310, 2001.

\bibitem{vimpactan2021}
Hao Tan, Jie Lei, Thomas Wolf, and Mohit Bansal.
\newblock {VIMPAC: Video Pre-Training via Masked Token Prediction and
  Contrastive Learning}, 2021.

\bibitem{Tran2015LearningSF}
Du Tran, Lubomir~D. Bourdev, Rob Fergus, Lorenzo Torresani, and Manohar Paluri.
\newblock {Learning Spatiotemporal Features with 3D Convolutional Networks}.
\newblock In {\em IEEE International Conference on Computer Vision}, 2015.

\bibitem{Tran2018ACloserLook}
Du Tran, Heng Wang, Lorenzo Torresani, Jamie Ray, Yann LeCun, and Manohar
  Paluri.
\newblock A closer look at spatiotemporal convolutions for action recognition.
\newblock In {\em IEEE Conference on Computer Vision and Pattern Recognition},
  2018.

\bibitem{KriegeskorteGoinginCircles}
Ruben~S {van Bergen} and Nikolaus Kriegeskorte.
\newblock Going in circles is the way forward: the role of recurrence in visual
  inference.
\newblock {\em Current Opinion in Neurobiology}, 65:176--193, 2020.
\newblock Whole-brain interactions between neural circuits.

\bibitem{TimesformerPytorch}
Heng Wang.
\newblock {TimeSformer-PyTorch. Implementation of TimeSformer from Facebook AI,
  a pure attention-based solution for video classification.}
\newblock \url{https://github.com/lucidrains/TimeSformer-pytorch}, 2021.
\newblock Accessed: 2021-11-13.

\bibitem{Wang2021TDNTD}
Limin Wang, Zhan Tong, Bin Ji, and Gangshan Wu.
\newblock Tdn: Temporal difference networks for efficient action recognition.
\newblock In {\em IEEE Conference on Computer Vision and Pattern Recognition},
  2021.

\bibitem{Wang2019TemporalSN}
Limin Wang, Yuanjun Xiong, Zhe Wang, Yu Qiao, Dahua Lin, Xiaoou Tang, and
  Luc~Van Gool.
\newblock Temporal segment networks for action recognition in videos.
\newblock {\em IEEE Transactions on Pattern Analysis and Machine Intelligence},
  41:2740--2755, 2019.

\bibitem{Xie2018RethinkingSF}
Saining Xie, Chen Sun, Jonathan Huang, Zhuowen Tu, and Kevin~P. Murphy.
\newblock Rethinking spatiotemporal feature learning: Speed-accuracy trade-offs
  in video classification.
\newblock In {\em European Conference on Computer Vision}, 2018.

\bibitem{VideoDGPAMI21}
Zhiyu Yao, Yunbo Wang, Jianmin Wang, Philip Yu, and Mingsheng Long.
\newblock {VideoDG: Generalizing Temporal Relations in Videos to Novel
  Domains}.
\newblock {\em IEEE Transactions on Pattern Analysis and Machine Intelligence},
  pages 1--1, 2021.

\bibitem{videocorruption2021}
Chenyu Yi, Siyuan Yang, Haoliang Li, Yap-Peng Tan, and Alex~C. Kot.
\newblock Benchmarking the robustness of spatial-temporal models against
  corruptions.
\newblock In {\em Advances in Neural Information Processing Systems}, 2021.

\bibitem{zhou2017temporalrelation}
Bolei Zhou, Alex Andonian, Aude Oliva, and Antonio Torralba.
\newblock Temporal relational reasoning in videos.
\newblock In {\em European Conference on Computer Vision}, 2018.

\end{thebibliography}
}

\appendix

\section{Supplemental figures regarding the model concepts}

\begin{figure*}[h!]
    \centering
 \includegraphics[width=0.9\textwidth]{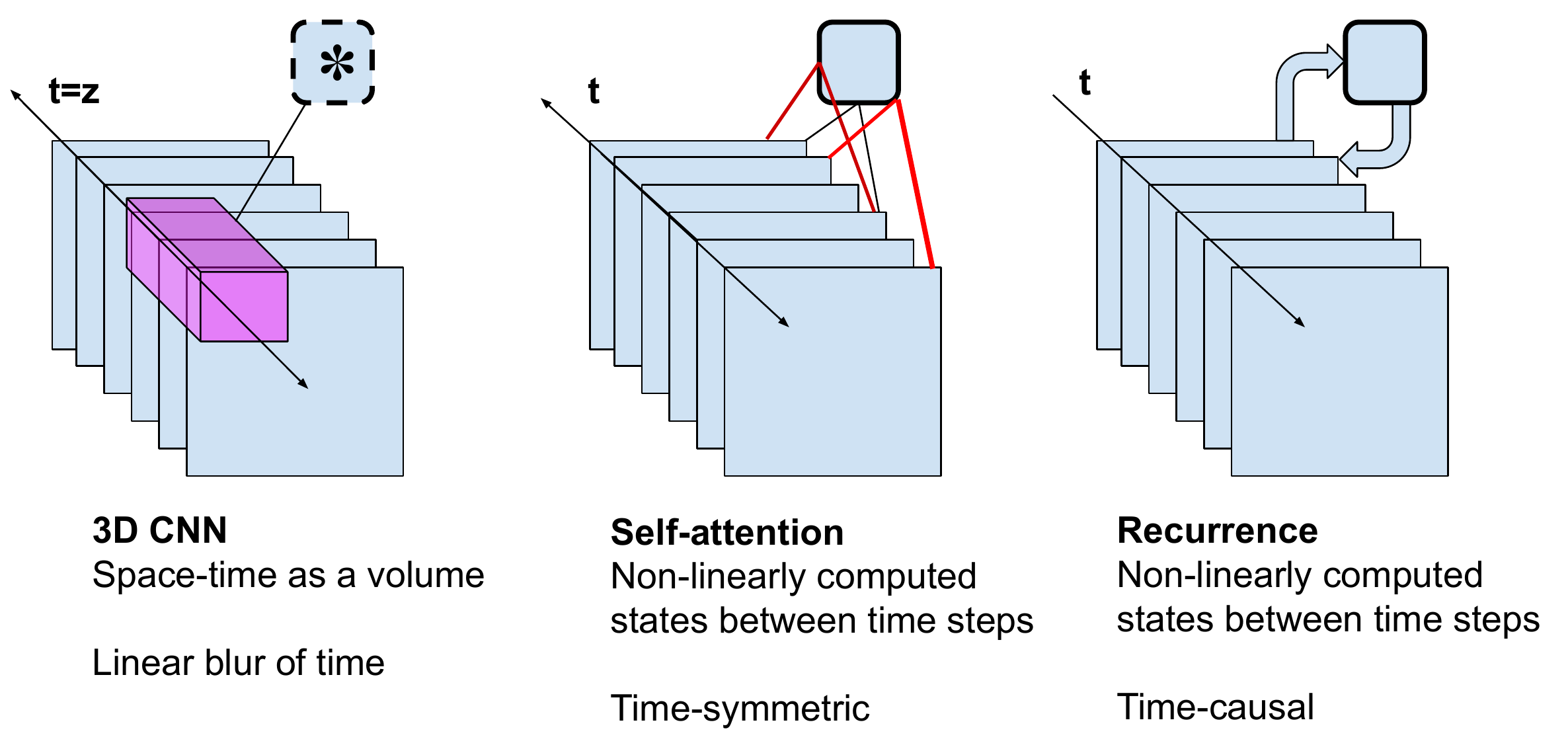}
    \label{fig:rac_overview}
    \caption{An overview of the conceptual differences in terms of frame dependency modeling between 3D convolutions, self-attention and recurrence.}
\label{fig:gradcam_clips}
\end{figure*}

Figure \ref{fig:rac_overview} highlights the conceptual differences between 3D convolution, self-attention and recurrence in terms of temporal modeling.

\section{Plots for each model size on the Temporal Shape dataset}

In the main article (Figure 3), the shaded area of standard error is across both model sizes and repeated runs with different seeds (meaning $10*5=50$ runs per model and domain). Detailed plots for each model size with five repeated runs each are shown in Figures \ref{fig:plot_drops_ten_2dot}-\ref{fig:plot_drops_ten_mnistbg}.

\begin{figure*}[]
\centering
\setlength{\figwidth}{.4\textwidth}
\setlength{\figheight}{.22\textheight}
\input{figures/plots/drops_ten_train2dot_ts}
\caption{Average results (\% acc.) across ten trials with varying numbers of hidden units per layer, repeated five times each. Training and validation on the 2Dot domain. The shaded area corresponds to standard deviation across the trials.
}
\label{fig:plot_drops_ten_2dot}
\end{figure*}

\begin{figure*}[]
\centering
\setlength{\figwidth}{.4\textwidth}
\setlength{\figheight}{.22\textheight}
\input{figures/plots/drops_ten_trainmnistbg_ts}
\caption{Average results (\% acc.) across ten trials with varying numbers of hidden units per layer, repeated five times each. Training and validation on the MNIST-bg domain. The shaded area corresponds to standard deviation across the trials.
}
\label{fig:plot_drops_ten_mnistbg}
\end{figure*}

\section{Robustness ratios for training both on 2Dot and MNIST-bg}

In the main article, robustness ratios vs. model size are only plotted when training on 2Dot. In Figure \ref{fig:plot_drop_vs_complexity_ts}, we include results when training on MNIST-bg as well. We show the two plots next to each other for comparison.

\begin{figure*}[t]
\centering
\setlength{\figwidth}{.34\textwidth}
\setlength{\figheight}{.20\textheight}
\input{figures/plots/rr_vsmodelcomplexity_2dotcol1_mnistbgcol2_withouttimesf1.tex}
\caption{Robustness ratio (rr.) ($\uparrow$)
when training on 2Dot (top -- same as in the main article for comparison) and on MNIST-bg (bottom), vs. number of hidden units per layer. The target domain is progressively further away from the source in subplots a-c. TimeSf-1 is excluded here due to its near random validation accuracy for small model sizes.}
\label{fig:plot_drop_vs_complexity_ts}
\end{figure*}

\section{Detailed results on Diving48}

Tables \ref{table:exp-e}-\ref{table:exp-h} show tabularized results corresponding to Figure 6 in the main article (experiments e-h).

\begin{table}[]
\scriptsize
\centering
\setlength{\tabcolsep}{9pt}
\begin{tabular}{lllll}
\textbf{Model}       & \textbf{S1/V} $\uparrow$ & \textbf{S2/V} $\uparrow$ & \textbf{T/V} $\downarrow$ & \textbf{T/S1} $\downarrow$ \\ \midrule

\textbf{3D CNN}      & $\mathbf{0.297}$  & $\mathbf{0.316}$  & 0.244             & $\mathbf{0.824}$ \\
\textbf{ConvLSTM}    & 0.245             & 0.279             & 0.238             & 0.973      \\
\textbf{TimeSformer} & 0.211             & 0.206             & $\mathbf{0.182}$  & 0.864    \\ \bottomrule                          
\end{tabular}
\caption{Results for experiment (e): 14M parameters, 30\% validation accuracy.}
\label{table:exp-e}
\end{table}

\begin{table}[]
\scriptsize
\centering
\setlength{\tabcolsep}{9pt}
\begin{tabular}{lllll}
\textbf{Model}       & \textbf{S1/V} $\uparrow$ & \textbf{S2/V} $\uparrow$ & \textbf{T/V} $\downarrow$ & \textbf{T/S1} $\downarrow$ \\ \midrule

\textbf{3D CNN}      & $\mathbf{0.279}$  & $\mathbf{0.303}$  & 0.216             & $\mathbf{0.777}$ \\
\textbf{ConvLSTM}    & 0.216             & 0.270             & 0.199             & 0.919      \\
\textbf{TimeSformer} & 0.155             & 0.162             & $\mathbf{0.150}$  & 0.972    \\ \bottomrule                          
\end{tabular}
\caption{Results for experiment (f): 14M parameters, 35\% validation accuracy.}
\label{table:exp-f}
\end{table}

\begin{table}[]
\scriptsize
\centering
\setlength{\tabcolsep}{9pt}
\begin{tabular}{lllll}
\textbf{Model}       & \textbf{S1/V} $\uparrow$ & \textbf{S2/V} $\uparrow$ & \textbf{T/V} $\downarrow$ & \textbf{T/S1} $\downarrow$ \\ \midrule

\textbf{3D CNN}      & $\mathbf{0.259}$  & $\mathbf{0.263}$  & 0.191             & $\mathbf{0.737}$ \\
\textbf{ConvLSTM}    & 0.183             & 0.224             & $\mathbf{0.142}$             & 0.776       \\ \bottomrule                          
\end{tabular}
\caption{Results for experiment (g): 14M parameters, 40\% validation accuracy.}
\label{table:exp-g}
\end{table}

\begin{table}[]
\scriptsize
\centering
\setlength{\tabcolsep}{9pt}
\begin{tabular}{lllll}
\textbf{Model}       & \textbf{S1/V} $\uparrow$ & \textbf{S2/V} $\uparrow$ & \textbf{T/V} $\downarrow$ & \textbf{T/S1} $\downarrow$ \\ \midrule

\textbf{3D CNN}      & $\mathbf{0.267}$  & $\mathbf{0.258}$  & 0.190             & $0.713$ \\
\textbf{ConvLSTM}    & 0.219             & 0.245             & $\mathbf{0.143}$  & $\mathbf{0.653}$       \\ \bottomrule                          
\end{tabular}
\caption{Results for experiment (h): 14M parameters, 45\% validation accuracy.}
\label{table:exp-h}
\end{table}

\section{Qualitative examples on Diving48}

Here, we include the top-1 and top-5 accuracies tables corresponding to the qualitative examples of classes 12, 22 and 45 shown in Table 7 in the main article. In Tables \ref{tab:qual12}, \ref{tab:qual22} and \ref{tab:qual45}, the trends regarding the top-1 and top-5 accuracy on the different datasets are slightly less clear. We observe that in Tables \ref{tab:qual12} and \ref{tab:qual45}, ConvLSTM and TimeSf drop the clearest in top-5 performance on T relative to S1 and S2. On the other hand, in Table \ref{tab:qual22} (Class 22), the top-5 accuracy is relatively improved on T  compared to S1 and S2 for ConvLSTM and the 3D CNN, whereas TimeSf is unchanged. We inspected these clips, to verify that the segmentation had not failed, which it had not. However, the ConvLSTM is still the only one out of the three to have 20\% in top-1 accuracy both for S1 and S2 on class 22, dropping to 0 in top-1 on T (Table \ref{tab:qual22}). Last, for class 45, the ConvLSTM has the best results on S1 and S2 (20\% top-5 accuracy) out of the three models, where the others have 0\% accuracy, except for 20\% top-5 accuracy for the 3D CNN on the texture dataset.

\begin{table}[ht]
\scriptsize
\centering
\setlength{\tabcolsep}{7pt}
\begin{tabular}{l|ll|ll|ll}
         & \multicolumn{2}{l}{\textbf{S1}} & \multicolumn{2}{l}{\textbf{S2}} & \multicolumn{2}{l}{\textbf{T}} \\ \toprule
\textbf{Model}& Top-1   & Top-5    & Top-1   & Top-5   & Top-1  & Top-5   \\ \midrule
ConvLSTM & 0.0     & 0.2  &  0.0    & 0.2     & 0.0    & 0.0    \\
3D CNN   & 0.2     & 0.2  & 0.2    & 0.2    & 0.2   & 0.2   \\
TimeSf   & 0.2    & 0.6   & 0.2     & 0.4    & 0.0   &  0.2    \\ \bottomrule
\end{tabular}
\caption{Qualitative example with predictions on five random clips from class 12, made by the model instances from experiment c) (38.3\% acc.).}
\label{tab:qual12}
\end{table}

\begin{table}[ht]
\scriptsize
\centering
\setlength{\tabcolsep}{7pt}
\begin{tabular}{l|ll|ll|ll}
         & \multicolumn{2}{l}{\textbf{S1}} & \multicolumn{2}{l}{\textbf{S2}} & \multicolumn{2}{l}{\textbf{T}} \\ \toprule
\textbf{Model}& Top-1   & Top-5  & Top-1   & Top-5  & Top-1  & Top-5  \\ \midrule
ConvLSTM & 0.2    & 0.2     & 0.2     & 0.2    & 0.0   & 0.6   \\
3D CNN   & 0.0    & 0.4    & 0.2    &    0.6   & 0.0   & 0.8      \\
TimeSf   & 0.0    & 0.2    & 0.0     & 0.2    & 0.0   & 0.2     \\ \bottomrule
\end{tabular}
\caption{Class 22, same table structure as Table \ref{tab:qual12}.}
\label{tab:qual22}
\end{table}

\begin{table}[ht]
\scriptsize
\centering
\setlength{\tabcolsep}{7pt}
\begin{tabular}{l|ll|ll|ll}
         & \multicolumn{2}{l}{\textbf{S1}} & \multicolumn{2}{l}{\textbf{S2}} & \multicolumn{2}{l}{\textbf{T}} \\ \toprule
\textbf{Model}& Top-1   & Top-5    & Top-1   & Top-5    & Top-1  & Top-5    \\ \midrule
ConvLSTM & 0.0    & 0.2     & 0.0    & 0.2    & 0.0   &  0.0     \\
3D CNN   & 0.0    & 0.2    & 0.0     & 0.0      & 0.0    &  0.2   \\
TimeSf   & 0.0    & 0.0    & 0.0     & 0.0     & 0.0   & 0.0    \\ \bottomrule
\end{tabular}
\caption{Class 45, same table structure as Table \ref{tab:qual12}.}
\label{tab:qual45}
\end{table}

\section{Dataset details}

\subsection{Sampling with Replacement in the Temporal Shape Dataset}

In the experiments, 4000 clips were used for training 
and 1000 for validation. The number of samples was chosen so as to be able to sample randomly with replacement, while still keeping the risk low 
that an identical clip occurs in both the training and the validation set. For the 2Dot-domain, each class has more than 30k possible variations (lower bounds: 31k circle, 34k line, 51k rectangle, 150k arc), except the spiral class which has 7200 as a lower bound on the possible variations. When the training set consists of 5000 samples in total, we generate around 1000 samples per class. 
For the spiral class, a frequentist estimation gives that $800/7200 = 0.11$ of the 200 spiral validation samples might be present in the training split (22 clips). However, this is still an over-estimation, since the spirals sometimes bounce against the sides of the frame which gives rise to extra variation. We decided to consider this as acceptable noise of the dataset. Some amount of data leakage can be considered interesting since this may occur in standard datasets as well. 

\subsection{Instance Segmentation of Diving48}

To segment divers, it did not suffice to apply a pre-trained network and use the class "Person", which we first attempted (DeeplabV3 pre-trained on MS-COCO, provided by PyTorch). First of all, the off-the-shelf model could often not recognize the divers in the air as the "Person" class -- they can be up-side down, or assume strange shapes in the air. Secondly, the model would often detect pixels of the "Person" class in the audience, when there was audience visible, which we, naturally, did not want to include.

Thus, we resorted to labelling our own segmented frames from the dataset (no segmentation masks were available online). We manually labelled 303 frames from the dataset containing one or two divers, picked from 303 randomly chosen videos of the training split. 
When there were two divers, we segmented each as its own instance. The segmentation masks will be made public. 

We fine-tuned a MaskRCNN on our labeled dataset, using a random split of 290 frames as training set and 13 frames to validate, and monitored the bounding box IoU on the validation set. The best model achieved 93\% validation bounding box IoU, which we used to segment the frames of the entire dataset (at 32 frames per clip). We used the confidence of the mask predictions as a threshold. The non-zero predictions were mostly confined to a bounding box surrounding the diver(s).
When the threshold was $t=0$, bounding boxes around the divers were used as crops (S2). When increased to $t=0.4$, we obtained proper segments of the diver shape (S1). The frames contain a lot of motion blur which made the segmentation more challenging, and the segmentation at $t=0.4$ is not perfect -- sometimes parts of for example an arm or foot is missing. 
 The performance of the segmentation at $t=0.4$ was deemed sufficient after manual inspection of 100 randomly chosen videos, where all videos had enough evidence to recognize the development of the dive. The segmentation at $t=0$ (bounding boxes, S2) was satisfactory in all 100 clips inspected.

\section{Parameter count}
Table \ref{tab:params} shows the number of parameter for the various architectures used in the Temporal Shape experiments.

%
%

\begin{table*}[]
\scriptsize
\caption{List of the number of trainable parameters for each model at each of the ten expreriments on Temporal Shape, where the model complexity was increased (the number of hidden units per layer, for three-layer models). TimeSformer-8 and TimeSformer-1 designates $\mathcal{A}=8$ or $\mathcal{A}=1$, respectively, i.e., the number of attention heads per layer.}
\centering
\begin{tabular}{@{}lllll@{}}
                             & \multicolumn{4}{l}{Nb. parameters}                                                    \\
                             \toprule
\textbf{\# hidden per layer} & \textbf{3D CNN} & \textbf{ConvLSTM} & \textbf{TimeSformer-8} & \textbf{TimeSformer-1} \\ \midrule
2                            & 1573            & 1497              & 20451                  & 877                    \\
4                            & 3573            & 4429              & 71557                  & 2229                     \\
6                            & 6005            & 8801              & 153413                 & 4061                     \\
8                            & 8869            & 14613             & 265989                 & 6373                   \\
12                           & 15893           & 30557             & 583301                 & 12437                  \\
16                           & 24645           & 52261             & 1023493                & 20421                  \\
24                           & 47333           & 112949            & 2272517                & 42149                  \\
32                           & 76933           & 196677            & 4013061                & 71557                  \\
48                           & 156869          & 433253            & 8968709                & 153413  \\ \bottomrule              
\end{tabular}
\label{tab:params}
\end{table*}

\begin{table*}[]
\tiny
\caption{List of the model variants used in the experiments a-h for Diving48. For the 3D CNN and ConvLSTM, the [x,y,z] lists designate the number of hidden units per layer (x for the first layer, y for the second, z for the third, etc), and the filter sizes lists similarly correspond to the filter size per layer.}
\centering
\begin{tabular}{@{}llll@{}}
                        &                                                                                                                                                                   &                                                                                                                          &                                                     \\
\textbf{Experiment}     & \textbf{3D CNN}                                                                                                                                                   & \textbf{ConvLSTM}                                                                                                        & \textbf{TimeSformer}                                \\ \midrule
\multicolumn{1}{|l|}{a} & \multicolumn{1}{l|}{\begin{tabular}[c]{@{}l@{}}Hidden {[}128,128,128,128{]}, \\ Filter sizes {[}7,7,5,3{]}\end{tabular}}                                          & \multicolumn{1}{l|}{\begin{tabular}[c]{@{}l@{}}Hidden {[}128,128,128,128{]},\\ Filter sizes {[}7,7,5,3{]}\end{tabular}}  & \multicolumn{1}{l|}{Depth=4, $D=1024$, $D_h = 128$} \\ \midrule
\multicolumn{1}{|l|}{b} & \multicolumn{1}{l|}{\begin{tabular}[c]{@{}l@{}}Hidden {[}128,128,128,128{]},\\ Filter sizes {[}7,7,5,3{]}\end{tabular}}                                           & \multicolumn{1}{l|}{\begin{tabular}[c]{@{}l@{}}Hidden {[}128,128,128,128{]}, \\ Filter sizes {[}7,7,5,3{]}\end{tabular}} & \multicolumn{1}{l|}{Depth=4, $D=1024$, $D_h = 128$} \\ \midrule
\multicolumn{1}{|l|}{c} & \multicolumn{1}{l|}{\begin{tabular}[c]{@{}l@{}}Hidden {[}128,128,128,128{]},\\ Filter sizes {[}7,7,5,3{]}\end{tabular}}                                           & \multicolumn{1}{l|}{\begin{tabular}[c]{@{}l@{}}Hidden {[}128,128,128,128{]},\\ Filter sizes {[}7,7,5,3{]}\end{tabular}}  & \multicolumn{1}{l|}{Depth=4, $D=1024$, $D_h = 128$} \\ \midrule
\multicolumn{1}{|l|}{d} & \multicolumn{1}{l|}{\begin{tabular}[c]{@{}l@{}}Hidden {[}32,64,128,128,128,256,\\ 256,256,512,512,512{]}\\ Filter sizes {[}5,3,3,3,3,3,3,3,3,3,3{]}\end{tabular}} & \multicolumn{1}{l|}{\begin{tabular}[c]{@{}l@{}}Hidden {[}128,128,128,128{]},\\ Filter sizes {[}7,7,5,3{]}\end{tabular}}  & \multicolumn{1}{l|}{-}                              \\ \midrule
\multicolumn{1}{|l|}{e} & \multicolumn{1}{l|}{\begin{tabular}[c]{@{}l@{}}Hidden {[}128,128,128,128,128,128{]},\\ Filter sizes {[}7,7,7,5,3,3{]}\end{tabular}}                               & \multicolumn{1}{l|}{\begin{tabular}[c]{@{}l@{}}Hidden {[}128,128,128,128{]},\\ Filter sizes {[}7,7,5,3{]}\end{tabular}}  & \multicolumn{1}{l|}{Depth=11, $D=256$, $D_h = 32$}  \\ \midrule
\multicolumn{1}{|l|}{f} & \multicolumn{1}{l|}{\begin{tabular}[c]{@{}l@{}}Hidden {[}128,128,128,128,128,128{]},\\ Filter sizes {[}7,7,7,5,3,3{]}\end{tabular}}                               & \multicolumn{1}{l|}{\begin{tabular}[c]{@{}l@{}}Hidden {[}128,128,128,128{]},\\ Filter sizes {[}7,7,5,3{]}\end{tabular}}  & \multicolumn{1}{l|}{Depth=11, $D=256$, $D_h = 32$}  \\ \midrule
\multicolumn{1}{|l|}{g} & \multicolumn{1}{l|}{\begin{tabular}[c]{@{}l@{}}Hidden {[}128,128,128,128,128,128{]},\\ Filter sizes {[}7,7,7,5,3,3{]}\end{tabular}}                               & \multicolumn{1}{l|}{\begin{tabular}[c]{@{}l@{}}Hidden {[}128,128,128,128{]},\\ Filter sizes {[}7,7,5,3{]}\end{tabular}}  & \multicolumn{1}{l|}{-}                              \\ \midrule
\multicolumn{1}{|l|}{h} & \multicolumn{1}{l|}{\begin{tabular}[c]{@{}l@{}}Hidden {[}128,128,128,128,128,128{]},\\ Filter sizes {[}7,7,7,5,3,3{]}\end{tabular}}                               & \multicolumn{1}{l|}{\begin{tabular}[c]{@{}l@{}}Hidden {[}128,128,128,128{]},\\ Filter sizes {[}7,7,5,3{]}\end{tabular}}  & \multicolumn{1}{l|}{-}                              \\ \bottomrule
\end{tabular}
\label{tab:divingmodelspecs}
\end{table*}

\section{TimeSformer variants attempted for training}

Table \ref{tab:timesfvariants} lists the different variants we tested when training on Diving48 from scratch. In all variants, the number of heads was 8 ($\mathcal{A} = 8$), the patch size was $16 \times 16$, the learning rate was fixed at 0.001, and the weight decay was 0.00001. When SGD was used, the momentum was always 0.9.

\begin{table*}[]
\scriptsize
\caption{List of attempted TimeSformer variants, trained from scratch on Diving48. $D$ and $D_h$ are parameters in the TimeSformer \cite{Timesformergberta_2021_ICML} 
architecture, attn. do. and ff.do are attention dropout and feed-forward network dropout, T is the number of uniformly sampled frames that constitute the clip, and addditional ll. means an additional linear layer on top of the predictions output from the TimeSformer model.}
\centering
\begin{tabular}{@{}llllllllllll@{}}
\toprule
Best val. & Ep. & $D$    & $D\_h$ & Depth & Attn. do. & Ff. do. & T  & Batch size & Optimizer & Additional ll. & Patience \\ \midrule
32.7      & 88  & 512  & 64   & 12    & 0             & 0           & 8  & 8          & SGD       & 1                       & 30       \\
31.5      & 84  & 512  & 64   & 12    & 0             & 0           & 8  & 8          & SGD       & 0                       & 30       \\
36.1      & 78  & 512  & 64   & 3     & 0             & 0           & 32 & 8          & SGD       & 0                       & 30       \\
39.7      & 122 & 1024 & 128  & 4     & 0             & 0           & 32 & 8          & SGD       & 0                       & 30       \\
31.1      & 76  & 512  & 64   & 12    & 0.1           & 0.1         & 8  & 8          & SGD       & 0                       & 30       \\
31.7      & 71  & 256  & 32   & 11    & 0             & 0           & 8  & 8          & SGD       & 0                       & 30       \\
36.5      & 85  & 256  & 32   & 11    & 0             & 0           & 32 & 8          & SGD       & 0                       & 30       \\
19.0      & 79  & 256  & 32   & 11    & 0             & 0           & 32 & 8          & Adam      & 0                       & 30       \\
31.7      & 75  & 256  & 32   & 11    & 0             & 0           & 8  & 32         & Adam      & 0                       & 30       \\
32.4      & 133 & 256  & 32   & 11    & 0             & 0           & 8  & 48         & SGD       & 0                       & 30       \\
36.5      & 85  & 256  & 32   & 11    & 0             & 0           & 32 & 8          & SGD       & 0                       & 75      \\ \bottomrule
\end{tabular}
\label{tab:timesfvariants}
\end{table*}

\section{Model specifications for the Diving48 experiments}

Table \ref{tab:divingmodelspecs} lists the different model specifications for each of the eight experiments a-h on Diving48 in the main article. For further details on the models, this is described in the main article and in the \href{https://github.com/sofiabroome/cross-dataset-generalization}{code repository}.

\end{document}